\documentclass[journal]{IEEEtran}
\usepackage{amsmath}
\usepackage{algorithm}
\usepackage{algorithmic}
\usepackage{graphicx}
\usepackage{multirow}
\usepackage{tabularx}
\usepackage{xcolor}
\usepackage{array}
\usepackage{amssymb}
\usepackage{bbding}
\usepackage{subfigure}
\usepackage{diagbox}
\usepackage{rotating}
\usepackage{booktabs}
\usepackage{cite}
\usepackage{CJK}
\usepackage{overpic}
\usepackage[colorlinks,
            linkcolor=red,
            anchorcolor=blue,
            citecolor=green
            ]{hyperref}

\newcommand{\addFig}[1]{}
\newcommand{\addFigs}[1]{}

\usepackage{subfigure}
\newcommand{\etal}{\textit{et~al}.~}
\newcommand{\ie}{\textit{i}.\textit{e}.,~}
\newcommand{\eg}{\textit{e}.\textit{g}.,~}

\usepackage{balance}
\usepackage{cleveref}
\crefformat{section}{\S#2#1#3} % see manual of cleveref, section 8.2.1
\crefformat{subsection}{\S#2#1#3}
\crefformat{subsubsection}{\S#2#1#3}

\ifCLASSINFOpdf
\else
\fi

\begin{document}
\title{Adjacent Context Coordination Network for Salient Object Detection in Optical Remote Sensing Images}

\author{Gongyang~Li,
	Zhi~Liu,~\IEEEmembership{Senior Member,~IEEE},
	Dan~Zeng,~\IEEEmembership{Senior Member,~IEEE}\\
	Weisi~Lin,~\IEEEmembership{Fellow,~IEEE},
        and~Haibin~Ling,~\IEEEmembership{Senior Member,~IEEE}
\thanks{Gongyang Li, Zhi Liu, and Dan Zeng are with Shanghai Institute for Advanced Communication and Data Science, Shanghai University, Shanghai 200444, China, and School of Communication and Information Engineering, Shanghai University, Shanghai 200444, China. Dan Zeng is also with the Key Laboratory of Specialty Fiber Optics and Optical Access Networks, Joint International Research Laboratory of Specialty Fiber Optics and Advanced Communication, Shanghai University, Shanghai 200444, China (email: ligongyang@shu.edu.cn; liuzhisjtu@163.com; dzeng@shu.edu.cn).}
\thanks{Weisi Lin is with the School of Computer Science and Engineering, Nanyang Technological University, Singapore 639798 (e-mail: wslin@ntu.edu.sg).}
\thanks{Haibin Ling is with the Department of Computer Science, Stony Brook University, Stony Brook, NY 11794 USA (email: hling@cs.stonybrook.edu).}% <-this % stops a space
\thanks{\textit{Corresponding author: Zhi Liu.}}
}

\markboth{IEEE TRANSACTIONS ON CYBERNETICS}%
{Shell \MakeLowercase{\textit{et al.}}: Bare Demo of IEEEtran.cls for IEEE Journals}

\maketitle

\begin{abstract}
Salient object detection (SOD) in optical remote sensing images (RSIs), or \textit{RSI-SOD}, is an emerging topic in understanding optical RSIs.
However, due to the difference between optical RSIs and natural scene images (NSIs), directly applying NSI-SOD methods to optical RSIs fails to achieve satisfactory results.
In this paper, we propose a novel Adjacent Context Coordination Network (ACCoNet) to explore the coordination of adjacent features in an encoder-decoder architecture for RSI-SOD.
Specifically, ACCoNet consists of three parts: an encoder, Adjacent Context Coordination Modules (ACCoMs), and a decoder.
As the key component of ACCoNet, ACCoM activates the salient regions of output features of the encoder and transmits them to the decoder.
ACCoM contains a local branch and two adjacent branches to coordinate the multi-level features simultaneously.
The local branch highlights the salient regions in an adaptive way, while the adjacent branches introduce global information of adjacent levels to enhance salient regions.
Additionally, to extend the capabilities of the classic decoder block (\ie several cascaded convolutional layers), we extend it with two bifurcations and propose a Bifurcation-Aggregation Block to capture the contextual information in the decoder.
Extensive experiments on two benchmark datasets demonstrate that the proposed ACCoNet outperforms 22 state-of-the-art methods under nine evaluation metrics, and runs up to 81 \textit{fps} on a single NVIDIA Titan X GPU.
The code and results of our method are available at https://github.com/MathLee/ACCoNet.
\end{abstract}

% Note that keywords are not normally used for peer review papers.

\IEEEpeerreviewmaketitle

\section{Introduction}
\IEEEPARstart{S}{alient} object detection (SOD) aims at distinguishing and highlighting visually attractive objects/regions in a scene, which has been extended from natural scene images~(NSIs)~\cite{2015SODBenchmark,Borji2019,2019sodsurvey} to videos~\cite{WWG19Video}, image groups~\cite{19CRMCO}, RGB-D images~\cite{Fan2019D3Net}, \textit{etc}.
It has many applications, such as object segmentation~\cite{LGY2019,LGY2021PFOS}, object tracking~\cite{15TRACK,20TRACK}, quality assessment~\cite{16SODIQA,19SGDNet}, hyperspectral image classification~\cite{2016SBS}, \textit{etc}.
Recently, SOD has been extended to optical remote sensing images (RSIs)~\cite{2015CIC,2015SSD,2018SPSS,2019SMFF,2019LVNet,2020PDFNet,2021DAFNet,2021PSL,2021MCCNet}, and has produced encouraging results.
For conciseness, in the rest of the paper, we use \textit{RSI-SOD} for the task of SOD in optical RSIs.

%%%
%%%
\begin{figure}[t!]
  \centering
  \begin{overpic}[width=1\columnwidth]{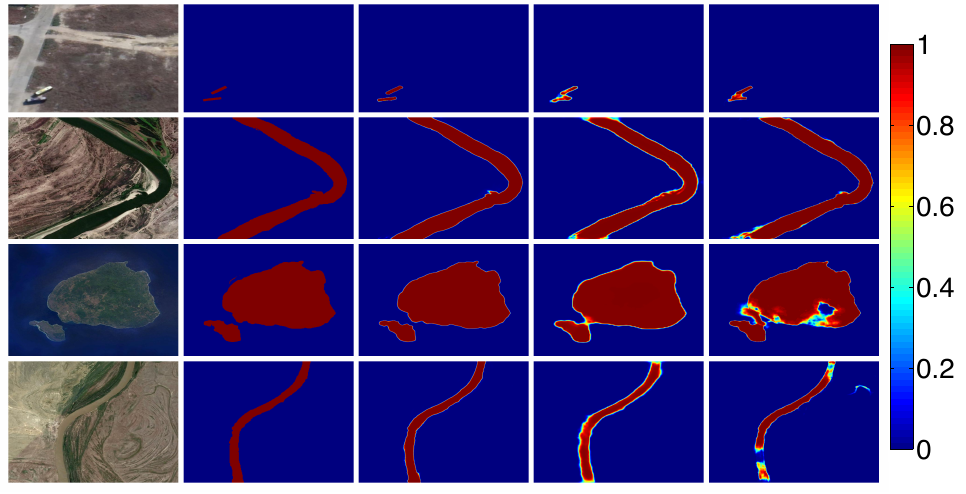}
    \put(1.5,-1.8){  \scriptsize{Optical RSI}  }
    \put(19.15,-1.8){ \scriptsize{Ground truth} }
    \put(41.4,-1.8){ \scriptsize{\textbf{Ours}} }
    \put(58.2,-1.8){ \scriptsize{DAFNet} }
    \put(75.9,-1.8){ \scriptsize{GateNet} }
  \end{overpic}
  \caption{Saliency maps produced by our method and two state-of-the-art methods DAFNet~\cite{2021DAFNet} and GateNet~\cite{2020GateNet} on optical RSIs.
  }\label{fig:example}
\end{figure}
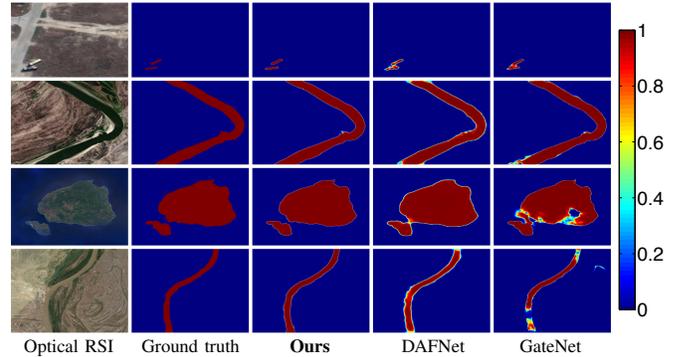
%%%
%%%

%
During the past decades, SOD in NSIs~\cite{2015SODBenchmark,Borji2019,2019sodsurvey}, or \textit{NSI-SOD} in short, has made a remarkable progress, especially when armed with deep learning techniques such as convolutional neural network (CNN)~\cite{1989CNN}.
Naturally, researchers will consider applying the mature NSI-SOD solutions to RSI-SOD.
However, there are significant differences in shooting devices, scenes and view orientations between NSIs and optical RSIs, resulting in differences in their resolutions, object types and object scales~\cite{2019LVNet,2021DAFNet}.
Consequently, direct migration of NSI-SOD solutions to RSI-SOD often leads to unsatisfactory performance.
As shown in the last column of Fig.~\ref{fig:example}, GateNet~\cite{2020GateNet}, which is a representative CNN-based NSI-SOD method and retrained on optical RSIs, cannot highlight salient objects in optical RSIs completely.
The existing specialized methods for RSI-SOD can be divided into traditional methods and CNN-based methods.
Traditional methods rely heavily on specific handcrafted features based on classical principles,
such as color information content~\cite{2015CIC}, sparse representation~\cite{2015SSD}, saliency feature analysis~\cite{2018SPSS}, and self-adaptive multiple feature fusion~\cite{2019SMFF}.
They usually fail in complex scenes of optical RSIs.
CNN-based methods focus on exploring effective feature interaction strategies to overcome the complex topology and unique scenes of optical RSIs.
The nested network~\cite{2019LVNet} fuses multi-resolution features;
the parallel down-up fusion network~\cite{2020PDFNet} focuses on the cross-path interaction, which is from low-level path/features to high-level path/features, between two adjacent features;
and the dense attention fluid network~(DAFNet)~\cite{2021DAFNet} transfers shallow-layer attention cues of low-level features, which capture edge and texture information, to deep layers, \ie high-level features, which capture semantic and object location information.
However, the influence of high-level features on low-level features is ignored, the coverage in feature interaction is insufficient, and the cascade structure of decoder blocks is plain, which may lead to incomplete exploration of the contextual information in optical RSIs.
As shown in the penultimate column of Fig.~\ref{fig:example}, the saliency maps of DAFNet~\cite{2021DAFNet}, which is currently the best specialized method, lose sharp boundaries and finer details.

Inspired by the above observations, in this paper, we propose a novel specialized solution for RSI-SOD, namely \textit{Adjacent Context Coordination Network} (ACCoNet), which focuses on coordinating adjacent features and capturing contextual information to adapt to diverse object types and object sizes in optical RSIs.
Our key idea is to comprehensively explore the contextual information contained in adjacent features, expand the coverage of feature interaction, and improve the context capture capability of plain decoder blocks.
Specifically, we consider features processing of three adjacent blocks (\ie the current, previous and subsequent blocks) in the backbone in a special module.
This way, the previous and subsequent features can provide comprehensive global auxiliary information to the current features.
Besides, we introduce bifurcations into plain decoder blocks to capture multi-scale context and increase the feature diversity.

In particular, we implement our ACCoNet in an encoder-decoder architecture.
ACCoNet is composed of an \emph{Adjacent Context Coordination Module} (ACCoM) for three adjacent features and a \emph{Bifurcation-Aggregation Block} (BAB) for the decoder.
ACCoM consists of three branches, one for local information and the other two for adjacent context.
Specifically, the local branch is responsible for modulating and enhancing current features in an adaptive manner,
while the other two adjacent branches are responsible for assisting current features with the previous and subsequent features through the previous-to-current and subsequent-to-current interactions.
For BAB, we put a bifurcation after each cascaded convolutional layer, and then aggregate these bifurcations to capture diverse contexts.
In this way, our ACCoNet achieves the best performance as compared with 22 state-of-the-art methods (an average $S_{\alpha}$ of 93.64\%, an average max $F_{\beta}$ of 89.93\% and an average max $E_{\xi}$ of 97.62\% on two datasets), and generates the most accurate saliency maps, as exemplified in the middle column of Fig.~\ref{fig:example}.

Our main contributions are summarized as follows:
\begin{itemize}
\item We explore the coordination of adjacent features in an encoder-decoder architecture for RSI-SOD, and propose a novel \emph{Adjacent Context Coordination Network} (ACCoNet), which effectively promotes the interaction of adjacent features for comprehensive coordination and fully captures contextual information, outperforming previous methods on public benchmarks.

\item We propose an \emph{Adjacent Context Coordination Module} (ACCoM) to coordinate cross-scale interactions in the feature embedding provided by the encoder and to deliver the valuable information to the decoder.

\item We extend the cascade structure of classic decoder blocks to the bifurcation-aggregation structure, and propose a \emph{Bifurcation-Aggregation Block} (BAB) to capture the multi-scale contextual information in the decoder.

\end{itemize}

The remaining parts of this paper are organized as follows.
In Sec.~\ref{sec:related}, we summarize the related works of NSI-SOD and RSI-SOD.
In Sec.~\ref{sec:OurMethod}, we elaborate our ACCoNet.
In Sec.~\ref{sec:exp}, we present the experiments and ablation studies of our ACCoNet.
Finally, the conclusion is drawn in Sec.~\ref{sec:con}.

\section{Related Work}
\label{sec:related}
In this section, we review the classic works of NSI-SOD and RSI-SOD, including traditional methods and CNN-based methods.

\subsection{Salient Object Detection in NSIs}
\label{sec:NSI_SOD}

\textit{1) Traditional NSI-SOD Methods.}
Salient object detection starts with natural scene images~\cite{1998Itti}, and a lot of traditional methods~\cite{2015SODBenchmark} have investigated hand-crafted features for NSI-SOD.
Traditional NSI-SOD methods can been divided into three categories: unsupervised methods~\cite{1998Itti,2014MERW,Liu2014ST,2015RRWR,2018RCRR,2015VPAA,2016HDCT,2017DSG,2017SMD,2019HSL}, semi-supervised methods~\cite{2019LFCS}, and supervised methods~\cite{2015FS}.
Numerous principles and technologies have been proposed for unsupervised methods,
such as center-surround differences~\cite{1998Itti},
the maximal entropy random walk~\cite{2014MERW},
the saliency tree~\cite{Liu2014ST},
the regularized random walks ranking~\cite{2015RRWR,2018RCRR},
directional information~\cite{2015VPAA},
the high-dimensional color transform~\cite{2016HDCT},
the sparse graph~\cite{2017DSG},
the structured matrix decomposition~\cite{2017SMD},
the hybrid sparse learning~\cite{2019HSL}, \textit{etc}.
Compared with unsupervised methods, there are relatively fewer semi-supervised and supervised methods in traditional methods.
Zhou~\etal\cite{2019LFCS} first utilized a boundary homogeneity model to generate pseudo labels.
Then based on a linear feedback control system model, they presented an iterative semi-supervised learning framework to establish relationships between control states and saliency map.
Liang~\etal\cite{2015FS} trained a support vector machine to select features through the supervised learning, which removes redundant features and speeds up model learning.
Wang~\etal\cite{2013SDMIL} presented a supervised multiple-instance learning framework for saliency detection, which incorporates a set of low-, mid-, and high-level features to comprehensively predict the scores of salient regions.

\textit{2) CNN-based NSI-SOD Methods.}
Different from traditional methods, most CNN-based NSI-SOD methods~\cite{Borji2019,2019sodsurvey} are based on supervised learning, and they greatly improve the detection accuracy.
A large number of well-known strategies of feature processing have been proposed, such as
the multi-level and multi-scale feature interaction~\cite{2019MIMF,2020MINet},
the feature suppress and balance~\cite{2020GateNet},
the sparse and dense labeling aggregation~\cite{2020DSDL},
the edge-aware feature fusion~\cite{2019EGNet,2020ITSD},
the global context-aware aggregation~\cite{2019PoolNet,2020GCPA}.
In addition, many popular mechanisms in the deep learning community are applied to NSI-SOD, such as
the deep supervision~\cite{2017DSS,2021DNA},
the recurrent mechanism~\cite{2018RADF,2018R3Net},
the attention mechanism~\cite{2019PFAN,2020GCPA,2020CAANet,2020EARNet},
the generative adversarial learning~\cite{2020ALN}, and
the adversarial attack~\cite{2020ROSA}.
Differently, Li \etal\cite{2020DNM} focused on the detection speed and proposed the depthwise nonlocal network, which achieves competitive performance using a single CPU thread.
Liu \etal\cite{2020HVPL} explored a lightweight architecture for NSI-SOD and imitated the primate visual cortex in their network via hierarchical visual perception learning.

The above NSI-SOD methods have a great influence on RSI-SOD methods.
However, due to the essential differences between NSIs and optical RSIs, these RSI-SOD methods have made specific modifications to the hand-crafted features or the CNN feature processing strategies of original NSI-SOD methods.

\subsection{Salient Object Detection in Optical RSIs}
\label{sec:ORSI_SOD}
As an emerging field, the SOD in optical RSIs, \ie RSI-SOD, has attracted more and more attention.
Zhang \etal\cite{2015CIC} first performed the color information content analysis on the input optical RSI to get the saliency scores of each color component, and then they constructed the saliency map based on these saliency scores.
Zhao \etal\cite{2015SSD} obtained low-level features via the global cues and background prior, and the sparse representation was introduced to transform low-level features to high-level features for saliency map integration.
In~\cite{2018SPSS}, Zhang \etal combined the super-pixel segmentation and statistical saliency feature analysis for RSI-SOD.
Zhang \etal\cite{2019SMFF} fused the features of color, intensity, texture and global contrast adaptively based on the low-rank matrix recovery to generate the saliency map.
Faur \etal\cite{2009RDM} combined the mean-shift-based segmentation and the rate distortion-based optimization together for salient remote sensing image segmentation.

Different from the above traditional RSI-SOD methods, CNN-based RSI-SOD solutions explore the unique characteristics from optical RSI data, and have made a promising progress.
Li \etal\cite{2019LVNet} constructed a challenging dataset for RSI-SOD.
They proposed an LV-shaped network, where the L-shaped two-stream pyramid module receives input images of five resolutions and the V-shaped nested connections structure infers salient objects based on multi-resolution features.
In~\cite{2020PDFNet}, Li \etal designed five parallel paths with dense connections, which exploit the in-path and cross-path information contained in two adjacent features to detect diversely scaled salient objects in optical RSIs.
Zhang \etal\cite{2021DAFNet} first established shallow-to-deep connections between different levels through dense attention fluid structure, and then they exploited global-context information to achieve feature alignment and reinforcement.
Zhang \etal\cite{2021PSL} combined the weakly and fully supervised learning for RSI-SOD.
They obtained pseudo annotations based on a classification network and the gradient-weighted class activation mapping to train the feedback saliency analysis network.
Tu \etal\cite{2021MJRBM} proposed a multiscale joint region and boundary model for RSI-SOD.
Following~\cite{2019LVNet}, Zhou \etal\cite{2021EMFINet} proposed a three inputs based edge-aware feature integration network.

Aside from the above studies, there are some works on tasks related to RSI-SOD,
such as airport detection~\cite{2018VOS},
building extraction~\cite{2017ISC},
residential areas extraction~\cite{2016GLSA},
ship detection~\cite{2019FHD},
oil tank detection~\cite{2019CMC,2019SGSM},
and region-of-interest detection/extraction~\cite{2014FDA,2016SPS,2017SDBD,2019SD}.
These methods show good performance in specific scenes of optical RSIs, but may not generalize well to various optical RSI scenes, resulting in poor performance in RSI-SOD.

As we know, the salient objects in optical RSIs usually have complex geometry structures, variable sizes and uncertain quantities, and are often accompanied with occlusion, shadows and abnormal illumination.
The specialized methods mentioned above put forward meaningful solutions to the characteristics of optical RSIs.
However, we believe that the contextual information in optical RSIs needs to be further explored, which is important to overcome these challenging scenes.
We thoroughly explore the contextual information in both encoder and decoder of our ACCoNet.
Concretely, the previous-to-current and subsequent-to-current feature interactions are established among three adjacent blocks in the encoder, and the cascade structure is updated to the bifurcation-aggregation structure in the decoder.

\section{Methodology}
\label{sec:OurMethod}
In this section, we elaborate the proposed Adjacent Context Coordination Network (ACCoNet).
In Sec.~\ref{sec:Overview}, we clarify the network overview and motivation of our ACCoNet.
In Sec.~\ref{sec:ACCM}, we present our Adjacent Context Coordination Module (ACCoM) in detail.
In Sec.~\ref{sec:BAB}, we give the detailed formulas of our Bifurcation-Aggregation Block (BAB).
In Sec.~\ref{sec:Loss Function}, we introduce the loss function.

\subsection{Network Overview and Motivation}
\label{sec:Overview}
\begin{figure*}
	\centering
	\begin{overpic}[width=1\textwidth]{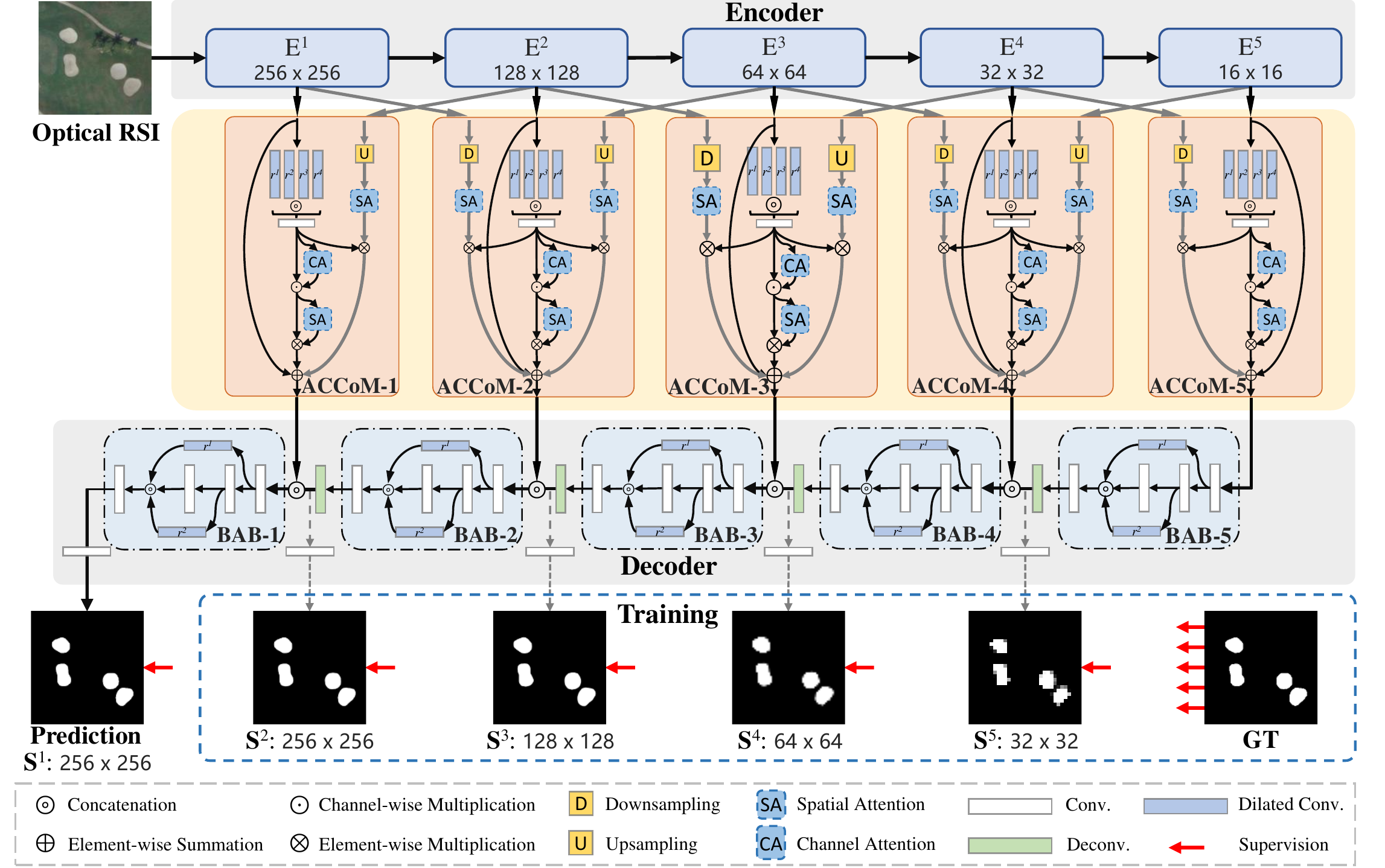}
    \end{overpic}
	\caption{Pipeline of the proposed ACCoNet, which is comprised of three key parts: the encoder network, the Adjacent Context Coordination Module (ACCoM) and the decoder network with Bifurcation-Aggregation Blocks (BABs).
	First, the encoder network extracts the basic features at five scales.
	Then, these basic features are fed to five ACCoMs to coordinate the feature activation.
	Finally, the output contextual features of ACCoM are transmitted to the decoder, which employs BABs to further capture contextual information, for inferring the salient objects.
	Notably, in the training phase, we adopt the deep supervision strategy, and attach the pixel-level supervision to each decoder block.
	GT denotes ground truth.
    }
    \label{fig:Framework}
\end{figure*}

The proposed ACCoNet is based on the encoder-decoder architecture, which has shown outstanding ability in pixel-level prediction tasks, such as semantic segmentation~\cite{17SegNet}, medical image segmentation~\cite{2015Unet}, NSI-SOD~\cite{2020GateNet,2019PoolNet} and RGB-D SOD~\cite{20ICNet,20CMWNet,21HAINet}.
As shown in Fig.~\ref{fig:Framework}, ACCoNet consists of an encoder network, several ACCoM components, and a decoder network with BABs.

\textit{1) Encoder Network.}
Following~\cite{17SegNet,20ICNet,20CMWNet,21HAINet}, we adopt the plain VGG-16~\cite{2014VGG16ICLR} as our basic encoder network, where the last max-pooling layer and three fully connected layers are truncated.
As shown at the top of Fig.~\ref{fig:Framework}, our encoder network consists of five blocks, denoted by E$^{t}$ ($\mathit{t}\in\{1, 2, 3, 4, 5\}$ is the block index),
and we adopt the feature map of the last convolutional layer of each block, \ie \textit{conv1-2}, \textit{conv2-2}, \textit{conv3-3}, \textit{conv4-3} and \textit{conv5-3}, denoted by $\boldsymbol{f}^{t}_\textrm{e} \in \mathbb{R}^{h_t\!\times\!w_t\!\times\!c_t}$ ($c_{t = \{ 1,2,3,4,5\}} = \{64,128,256,512,512\}$), for subsequent processing.
The input size of our encoder network is $256\times256\times3$, so $h_t = \frac{256}{2^{t-1}}$ and $w_t = \frac{256}{2^{t-1}}$.

%%%
%%%
%%%
%%%

%
\textit{2) Adjacent Context Coordination Module.}
Contextual information is crucial for RSI-SOD.
It exists not only in one feature level, but also in features at adjacent levels.
Using convolutional layers with different convolution kernels in parallel is a popular strategy to capture local and global contents within one feature level.
This is conducive to capturing salient objects with variable sizes or uncertain quantities  in optical RSIs.
And introducing feature interaction among features at adjacent levels is an effective strategy to capture cross-level contextual complementary information.
This is effective for refining the details and determining the location of salient objects in optical RSIs.
Thus motivated, we explore the above two kinds of contextual information with these two mentioned strategies.
Since high-level features provide a lot of semantic clues and low-level features provide a lot of fine details, we coordinate cross-scale features from the current, previous and subsequent blocks.

In practice, we design three branches (\ie one local branch and two adjacent branches) in ACCoM.
The local branch is based on the first strategy.
Moreover, it is equipped with the attention mechanism for further feature modulation in an adaptive way.
The two adjacent branches are based on the second strategy, and consist of the previous-to-current branch and the subsequent-to-current branch.
Since that the previous and subsequent features are different in scale from the current features, the two adjacent branches provide cross-scale information via two spatial attention maps to align salient regions twice.
Comprehensive coordination enables the proposed ACCoM to transmit valuable contextual information to the decoder.
Notably, as shown in Fig.~\ref{fig:Framework}, for ACCoM-1 and ACCoM-5, due to their special position, we can only make one adjacent branch in them.
We present ACCoM in detail in Sec.~\ref{sec:ACCM}, and assess its effectiveness in Sec.~\ref{Ablation Studies}.

\textit{3) Bifurcation-Aggregation Block.}
The decoder network is in charge of inferring the salient objects.
Generally, the classic decoder network~\cite{17SegNet,2015Unet} is comprised of five plain decoder blocks, in which the convolutional layers are cascaded.
However, the inference ability of the cascade structure depends more on the features transmitted by the encoder, and the cascade structure is not sensitive to the unique scenes of optical RSIs, which may damage the inference accuracy of salient objects of the decoder network.
As previously mentioned, contextual information is crucial for RSI-SOD, so we further explore them in the decoder.
We introduce dilated convolutions~\cite{Dila2016} as bifurcations after the first two cascaded convolutional layers, and then aggregate the information from the two bifurcations and the original one via the concatenation-convolution operation.
In this way, the bifurcation-aggregation structure enriches the topology of the decoder through two dilated convolutions, expands the receptive field of features and captures rich contextual information, which is beneficial for inferring the salient objects.
We present BAB in detail in Sec.~\ref{sec:BAB}, and show its ablation studies in Sec.~\ref{Ablation Studies}.

\subsection{Adjacent Context Coordination Module}
\label{sec:ACCM}
\textit{Adjacent Context Coordination Module} is the key component in ACCoNet.
It connects the encoder and the decoder, and its details are illustrated in Fig.~\ref{fig:Framework}. 
There are usually three branches in ACCoM (\eg ACCoM-2, ACCoM-3 and ACCoM-4): one local branch (the middle one in ACCoM) and two adjacent branches (the left and right ones in ACCoM).
While ACCoM-1 and ACCoM-5 only contain two branches: one local branch and one adjacent branch.
Thus, we  generally define the processing of ACCoM as $\mathrm{F} (\cdot)$, which is formulated as follows:
\begin{equation}
   \begin{aligned}
    \boldsymbol{f}^{t}_{\textrm{accom}}=\left\{
	\begin{array}{lll}
	\mathrm{F} ( \boldsymbol{f}^{t}_\textrm{e}, \boldsymbol{f}^{t+1}_\textrm{e} ) ,     & t=1\\
	\mathrm{F} ( \boldsymbol{f}^{t-1}_\textrm{e}, \boldsymbol{f}^{t}_\textrm{e} , \boldsymbol{f}^{t+1}_\textrm{e} ) ,      & t=2,3,4\\
	\mathrm{F} ( \boldsymbol{f}^{t-1}_\textrm{e}, \boldsymbol{f}^{t}_\textrm{e} ) ,      & t=5,\\
	\end{array}  \right. 
    \label{eq:1}
    \end{aligned}
\end{equation}
where $\boldsymbol{f}^{t}_{\textrm{accom}} \in \mathbb{R}^{h_t\!\times\!w_t\!\times\!c_t}$ is the output feature of ACCoM-\textit{t}, and
$\boldsymbol{f}^{t-1}_\textrm{e}$, $\boldsymbol{f}^{t}_\textrm{e}$ and $\boldsymbol{f}^{t+1}_\textrm{e}$ are the previous, current and subsequent features, respectively.

\textit{1) Local Branch.}
The local branch operates on the current features $\boldsymbol{f}^{t}_\textrm{e} \in \mathbb{R}^{h_t\!\times\!w_t\!\times\!c_t}$, and contains two main operations.
First, we apply four dilated convolutions~\cite{Dila2016} (rather than normal convolutional layers) with different dilation rates in parallel to $\boldsymbol{f}^{t}_\textrm{e}$, which is defined as follows:
\begin{equation}
   \begin{aligned}
   \boldsymbol{f}^{t,i}_\textrm{dc} = \mathrm{DConv}_{\sigma}(\boldsymbol{f}^{t}_\textrm{e};\mathbf{W}^{t,i}_{3\!\times\!3}, r^{i}), ~ i\in\{1, 2, 3, 4\},
    \label{eq:2}
    \end{aligned}
\end{equation}
where $ \boldsymbol{f}^{t,i}_\textrm{dc} \in \mathbb{R}^{h_t\!\times\!w_t\!\times\!c_t }$ is the output feature of each dilated convolution,
$\mathrm{DConv}_{\sigma}(\ast;\ast,\ast)$ is the dilated convolution with Batch Normalization (BN)~\cite{2015BN} and ReLU activation function $\sigma$,
$\mathbf{W}^{t,i}_{3\!\times\!3}$ is the parameters with $3\!\times\!3$ kernel, and $r^i=i$
%$r^{i=\{1, 2, 3, 4\}}= \{1, 2, 3, 4\}$
is the dilation rate.
This can effectively traverse regions of different sizes in $\boldsymbol{f}^{t}_\textrm{e}$.

Then, we summarize these output features using the concatenation-convolution operation, obtaining features with rich contextual cues, \ie $ \boldsymbol{f}^{t}_\textrm{c} \in \mathbb{R}^{h_t\!\times\!w_t\!\times\!c_t }$, which is defined as follows:
\begin{equation}
   \begin{aligned}
    \boldsymbol{f}^{t}_\textrm{c}=  \mathrm{Conv}_{\sigma} \big( \mathrm{Concat}\big(   \boldsymbol{f}^{t,1}_\textrm{dc}, \boldsymbol{f}^{t,2}_\textrm{dc},  \boldsymbol{f}^{t,3}_\textrm{dc}, \boldsymbol{f}^{t,4}_\textrm{dc} \big); \mathbf{W}^{t}_{3\!\times\!3} \big),
    \label{eq:3}
    \end{aligned}
\end{equation}
where $\mathrm{Concat}(\cdot)$ is the cross-channel concatenation,
and $\mathrm{Conv}_{\sigma}(\ast;\ast)$ is the normal convolutional layer with BN and ReLU activation function.
The subsequent operations in ACCoM are based on $ \boldsymbol{f}^{t}_\textrm{c} $.

However, the summary operation is relatively straightforward, resulting in some redundant information in $ \boldsymbol{f}^{t}_\textrm{c} $.
We adopt the subtle channel attention (CA) and spatial attention (SA)~\cite{SENet,2018CBAM} to further purify $ \boldsymbol{f}^{t}_\textrm{c} $ in an adaptive manner, which is formulated as follows:
\begin{equation}
   \begin{aligned}
    \boldsymbol{f}^{t}_\textrm{loc} =  \mathrm{SA}( \mathrm{CA}(\boldsymbol{f}^{t}_\textrm{c}) \odot \boldsymbol{f}^{t}_\textrm{c} ) \otimes \boldsymbol{f}^{t}_\textrm{c},
    \label{eq:4}
    \end{aligned}
\end{equation}
where $\boldsymbol{f}^{t}_\textrm{loc} \in \mathbb{R}^{h_t\!\times\!w_t\!\times\!c_t}$ is the output feature of the local branch,
$\odot$ is the channel-wise multiplication, and
$\otimes$ is the element-wise multiplication.
Specifically, we implement CA with a spatial-wise global max pooling (GMP), a fully connected layer with ReLU activation function and a fully connected layer with sigmoid activation function; and we implement SA with a channel-wise GMP and a convolutional layer with sigmoid activation function.
Such an adaptive modulation process selects valuable contents from $\boldsymbol{f}^{t}_\textrm{c}$.

%%%
%%%
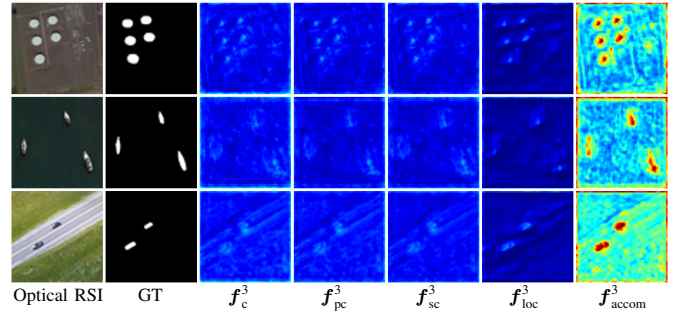
\begin{figure}
\centering
\footnotesize
  \begin{overpic}[width=1\columnwidth]{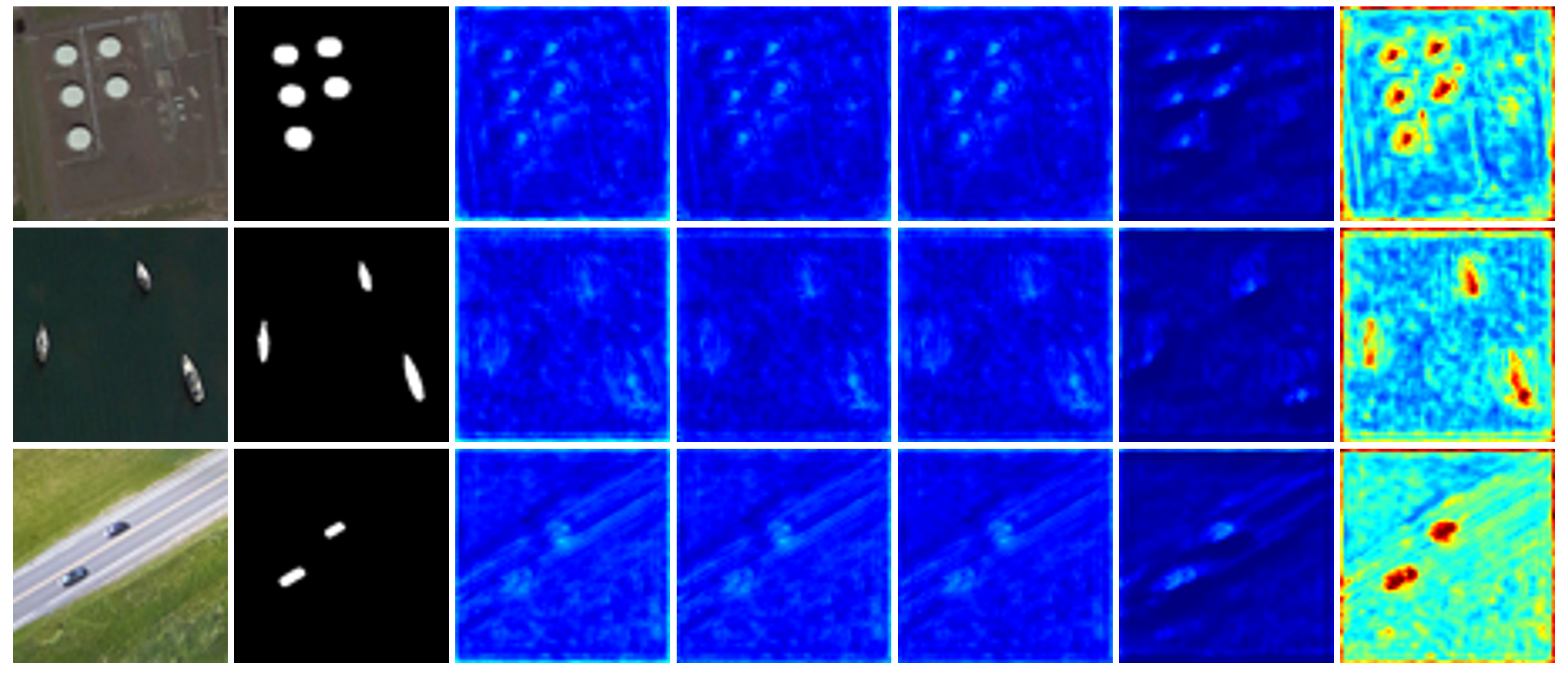}
    \put(-0.1,-2.1){ \scriptsize{Optical RSI} }
    \put(18.6,-2.1){ \scriptsize{GT} }
    \put(31.1,-2.3){ \scriptsize{ $\boldsymbol{f}^{\textrm{3}}_\textrm{c}$ } }
    \put(45.0,-2.3){ \scriptsize{ $\boldsymbol{f}^{\textrm{3}}_\textrm{pc}$  } }
    \put(59.2,-2.3){ \scriptsize{ $\boldsymbol{f}^{\textrm{3}}_\textrm{sc}$ }  }
    \put(73.2,-2.3){ \scriptsize{ $ \boldsymbol{f}^{\textrm{3}}_\textrm{loc} $ } }
    \put(86.6,-2.3){ \scriptsize{ $\boldsymbol{f}^{\textrm{3}}_{\textrm{accom}}$  }}
  \end{overpic}
\caption{
Feature visualization of each branch in ACCoM-3.
Please zoom-in for viewing details.
}
\label{feature_visual}
\end{figure}
%%%
%%%

%
\textit{2) Adjacent Branch(es).}
The adjacent branches contribute two types of assistance to $\boldsymbol{f}^{t}_\textrm{c}$.
The first one is the previous-to-current branch, which can be computed as:
\begin{equation}
   \begin{aligned}
    \boldsymbol{f}^{t}_\textrm{pc} = \mathrm{SA}( \mathrm{Down}(\boldsymbol{f}^{t-1}_\textrm{e}) ) \otimes \boldsymbol{f}^{t}_\textrm{c},  ~~~ t=2,3,4,5,
    \label{eq:5}
    \end{aligned}
\end{equation}
where $ \boldsymbol{f}^{t}_\textrm{pc} \in \mathbb{R}^{h_t\!\times\!w_t\!\times\!c_t}$ is the output feature of the previous-to-current branch, and $\mathrm{Down}(\cdot)$ is the 2$\times$ downsampling implemented by max-pooling.
This branch brings alignment information with fine details to $\boldsymbol{f}^{t}_\textrm{c}$.

The second one is the subsequent-to-current branch, which can be computed as:
\begin{equation}
   \begin{aligned}
    \boldsymbol{f}^{t}_\textrm{sc} = \mathrm{SA}( \mathrm{Up}(\boldsymbol{f}^{(t+1)}_\textrm{e}) )\otimes \boldsymbol{f}^{t}_\textrm{c},  ~~~ t=1,2,3,4,
    \label{eq:6}
    \end{aligned}
\end{equation}
where $ \boldsymbol{f}^{t}_\textrm{sc} \in \mathbb{R}^{h_t\!\times\!w_t\!\times\!c_t}$ is the output feature of the subsequent-to-current branch, and $\mathrm{Up}(\cdot)$ is the 2$\times$ upsampling implemented by bilinear interpolation.
This branch brings alignment information with object location to $\boldsymbol{f}^{t}_\textrm{c}$.

\textit{3) Branches Integration.}
After the above effective coordination, we integrate the output features of these three (or two) branches with the original current features as follows:
\begin{equation}
   \begin{aligned}
    \boldsymbol{f}^{t}_{\textrm{accom}} =\left\{
	\begin{array}{lll}
	\boldsymbol{f}^{t}_\textrm{loc}  \oplus \boldsymbol{f}^{t}_\textrm{sc}  \oplus \boldsymbol{f}^{t}_\textrm{e},     & t=1\\
	\boldsymbol{f}^{t}_\textrm{loc}  \oplus (\boldsymbol{f}^{t}_\textrm{pc}  \oplus \boldsymbol{f}^{t}_\textrm{sc} )  \oplus \boldsymbol{f}^{t}_\textrm{e} ,      & t=2,3,4\\
	\boldsymbol{f}^{t}_\textrm{loc}  \oplus  \boldsymbol{f}^{t}_\textrm{pc}  \oplus \boldsymbol{f}^{t}_\textrm{e} ,      & t=5,\\
	\end{array}  \right. 
    \label{eq:7}
    \end{aligned}
\end{equation}
where $\oplus$ is the element-wise summation and the original current features are regarded as the basic content.
In summary, $\boldsymbol{f}^{t}_\textrm{e}$ is coordinated by various contents, which greatly enhances the robustness and stability of $ \boldsymbol{f}^{t}_{\textrm{accom}}$.

In Fig.~\ref{feature_visual}, we visualize features in ACCoM-3.
It shows that, with all branches (\ie $\boldsymbol{f}^{\textrm{3}}_\textrm{loc}$, $\boldsymbol{f}^{\textrm{3}}_\textrm{pc}$ and $\boldsymbol{f}^{\textrm{3}}_\textrm{sc}$) working together, ACCoM accurately activates each salient region through comprehensive coordination, making the salient objects in $\boldsymbol{f}^{\textrm{3}}_{\textrm{accom}}$ more obvious than those in $\boldsymbol{f}^{\textrm{3}}_\textrm{c}$.

\subsection{Bifurcation-Aggregation Block}
\label{sec:BAB}
\textit{Bifurcation-Aggregation Block} is the basic unit of the decoder.
It processes the features from the current ACCoM and the previous BAB, and finally infers the mask of salient objects.
We define the processing of BAB as $\mathrm{B}(\cdot)$, which is formulated as follows:
\begin{equation}
   \begin{aligned}
    \boldsymbol{f}^{t}_{\textrm{bab}}=\left\{
	\begin{array}{lll}
	\mathrm{B} ( \boldsymbol{f}^{t}_{\textrm{accom}} , \mathrm{Deconv} ( \boldsymbol{f}^{t+1}_{\textrm{bab}} )) ,     & t=1,2,3,4\\
	\mathrm{B} ( \boldsymbol{f}^{t}_{\textrm{accom}}  ) ,      & t=5,\\
	\end{array}  \right. 
    \label{eq:8}
    \end{aligned}
\end{equation}
where $\boldsymbol{f}^{t}_{\textrm{bab}} $ is the output feature of BAB-\textit{t}, and $\mathrm{Deconv} (\cdot)$ is the deconvolution layer with BN and ReLU activation function.

For convenience, we define the features generated by the three cascaded convolutional layers in BAB-\textit{t} as $\boldsymbol{f}^{t,l}_{\textrm{b-c}}$ ($l \in \{1,2,3\}$).
So the output feature of two bifurcations (\ie $\boldsymbol{f}^{t,l}_{\textrm{bif}}$) can be computed as:
\begin{equation}
   \begin{aligned}
    \boldsymbol{f}^{t, l}_{\textrm{bif}}= \mathrm{DConv}_{\sigma}( \boldsymbol{f}^{t,l}_{\textrm{b-c}}; \mathbf{W}^{t,l}_{3\!\times\!3}, r^{l} ), ~~~ l = 1,2,
    \label{eq:9}
    \end{aligned}
\end{equation}
in which we adopt the dilated convolution to expand the receptive field and capture contextual cues from $\boldsymbol{f}^{t}_{\textrm{accom}}$.
In practice, considering the difference in feature resolution of each BAB, we set different dilation rates of bifurcations for different BABs.
The detailed parameters are shown in Tab.~\ref{tab:BAB_Details}.

%%%
%%%
\begin{table}[!t]
\centering
%\small
\caption{Detailed parameters of two bifurcations (\ie dilated convolutions) in BAB, including kernel size, channel number, dilation rate and the size of output feature.
  For instance, $(3\times3, 64, 64)$ denotes that the kernel size is $3\times3$, the input channel number is 64, and the output channel number is 64.
  }
\label{tab:BAB_Details}
%\footnotesize
%\setlength\tabcolsep{4pt}
\renewcommand{\arraystretch}{1.25}
\renewcommand{\tabcolsep}{2.8mm}
%\small
\begin{tabular}{c|c|c|c|c}
\bottomrule
  \hline
  Aspects & Dilated conv. & r$^\textrm{1}$ & r$^\textrm{2}$ & Output size\\
  \hline
  \hline
    BAB-1 & $(3\times3, 64, 64)$     & $5$ & $3$ & $[256\times256\times64]$  \\
    BAB-2 & $(3\times3, 128, 128)$ & $5$ & $3$ & $[128\times128\times128]$  \\
    BAB-3 & $(3\times3, 256, 256)$ & $5$ & $3$ & $[64\times64\times256]$  \\
    BAB-4 & $(3\times3, 512, 512)$ & $3$ & $2$ & $[32\times32\times512]$  \\
    BAB-5 & $(3\times3, 512, 512)$ & $3$ & $2$ & $[16\times16\times512]$  \\
\toprule
\end{tabular}
\end{table}
%%%
%%%

%
Then, we adopt the concatenation-convolution operation to aggregate these two bifurcations and the original $\boldsymbol{f}^{t,\textrm{3}}_{\textrm{b-c}}$ as:
\begin{equation}
   \begin{aligned}
    \boldsymbol{f}^{t}_{\textrm{bab}}= \mathrm{Conv}_{\sigma}( \mathrm{Concat} (\boldsymbol{f}^{t, \textrm{1}}_{\textrm{bif}}, \boldsymbol{f}^{t, \textrm{2}}_{\textrm{bif}}, \boldsymbol{f}^{t,\textrm{3}}_{\textrm{b-c}}); \mathbf{W}^{t}_{3\!\times\!3}).
    \label{eq:10}
    \end{aligned}
\end{equation}
This way, BAB further scans regions with different sizes based on $\boldsymbol{f}^{t}_{\textrm{accom}}$  at the inference stage, which can be well adapted to the characteristics of changes in the shape, size, and quantity of salient objects in optical RSIs.

\subsection{Loss Function}
\label{sec:Loss Function}
As shown at the bottom of Fig.~\ref{fig:Framework}, in the training phase, we attach the pixel-level supervision to each decoder block (\ie the deep supervision strategy~\cite{2015DeepSup}) for quick convergency.
Specifically, we arrange a convolutional layer after BAB-\textit{t} to generate the intermediate/final saliency map, denoted as $\mathbf{S}^{t}$.
For $\mathbf{S}^{t}$, we combine the pixel-level binary cross-entropy (BCE) loss with the map-level intersection-over-union (IoU) loss~\cite{19BASNet,21HAINet} for comprehensive and complementary content enhancement.
We formulate the total loss function $\mathbb{L}$ as:
\begin{equation}
   \begin{aligned}
    \mathbb{L}  = \sum\limits_{t=1}^5 \left( \mathrm{L}^{t}_\textrm{bce} (\mathrm{Up}(\mathbf{S}^{t}),\mathbf{GT}) + \mathrm{L}^{t}_\textrm{iou} (\mathrm{Up}(\mathbf{S}^{t}),\mathbf{GT}) \right),
    \label{eq:TotalLoss}
    \end{aligned}
\end{equation}
where $\mathrm{L}^{t}_\textrm{bce} (\cdot,\cdot)$ is the BCE loss, $\mathrm{L}^{t}_\textrm{iou} (\cdot,\cdot)$ is the IoU loss, and $\mathbf{GT}$ is the ground truth.
In this way, the deep supervision strategy with hybrid losses not only stabilizes our ACCoNet training process, but also improves the detection accuracy.

%%%
%%%
\begin{table*}[t!]
  \centering
  \caption{
    Quantitative comparison of our method and other 22 state-of-the-art methods,
    including five traditional NSI-SOD methods, ten CNN-based NSI-SOD methods, and seven RSI-SOD methods, on two popular datasets
  in terms of S-measure, maximum, mean and adaptive F-measure, maximum, mean and adaptive E-measure, and MAE.
  We also report the Frames Per Second (FPS) of all methods.
   $\uparrow$ and $\downarrow$ indicate larger and smaller is better, respectively.
    The top three results are marked in \textcolor{red}{\textbf{red}}, \textcolor{blue}{\textbf{blue}} and \textcolor{green}{\textbf{green}}, respectively.
    $^\dagger$ means deep learning based method.
    For simplicity, R3. is R3Net, Pool. is PoolNet, EG. is EGNet, MI. is MINet, Gate. is GateNet, LV. is LVNet, DAF. is DAFNet, MJRB. is MJRBM, EMFI. is EMFINet, and ACCo. is ACCoNet. 
    }
\label{table:QuantitativeResults}
  \footnotesize
  \renewcommand{\arraystretch}{1.25}
  \renewcommand{\tabcolsep}{0.10mm}
  \resizebox{1\textwidth}{!}{
  \begin{tabular}{lr|ccccc|cccccccccc|ccccccc|c}
  \hline
  \toprule
  
    &  & \multicolumn{5}{c|}{Traditional NSI-SOD Methods} 
    & \multicolumn{10}{c|}{CNN-based NSI-SOD Methods} 
    & \multicolumn{7}{c|}{RSI-SOD Methods} \\
    \cmidrule(l){3-7} \cmidrule(l){8-17} \cmidrule(l){18-24}
 
    &  & RRWR & HDCT & DSG & SMD & RCRR
    & DSS$^\dagger$ & RADF$^\dagger$ & R3.$^\dagger$ & PFAN$^\dagger$ & Pool.$^\dagger$ & EG.$^\dagger$
    & GCPA$^\dagger$ & MI.$^\dagger$ & ITSD$^\dagger$ & Gate.$^\dagger$ 
    & VOS & CMC & SMFF & LV.$^\dagger$ & DAF.$^\dagger$ & MJRB.$^\dagger$ & EMFI.$^\dagger$ & \textbf{ACCo.$^\dagger$} \\
    
    &  & 2015 & 2016 & 2017 & 2017 & 2018
    & 2017 & 2018 & 2018 & 2019 & 2019 & 2019 
    & 2020 & 2020 & 2020 & 2020
    & 2018 & 2019 & 2019 & 2019 & 2021 & 2022 & 2022 & 2022 \\
    
     & Metric & ~\cite{2015RRWR} & \cite{2016HDCT} & \cite{2017DSG} & \cite{2017SMD}  & \cite{2018RCRR} 
     & \cite{2017DSS} & \cite{2018RADF} & \cite{2018R3Net} & \cite{2019PFAN} & \cite{2019PoolNet} & \cite{2019EGNet}
     & \cite{2020GCPA} & \cite{2020MINet} & \cite{2020ITSD} & \cite{2020GateNet}   
     & \cite{2018VOS} & \cite{2019CMC} & \cite{2019SMFF} & \cite{2019LVNet} & \cite{2021DAFNet} & \cite{2021MJRBM} & \cite{2021EMFINet} & Ours\\

    \midrule   
%     & Time~$\downarrow$ & 2.910 & 7.130 & 1.570 & -- & 3.140 
%     & 0.120 & 0.150 & 0.480 & -- & -- & -- 
%     & 0.044 & 0.086 & 0.063 & 0.040
%     & -- & -- & -- & 0.740 & 0.038 & 0.012 \\  %0.01113
     
     & FPS~$\uparrow$ & 0.3 & 7 & 0.6 & -- & 0.3 
     & 8 & 7 & 2 & -- & -- & -- 
     & 23 &12 & 16 & 25
     & -- & -- & -- & 1.4 & 26 & 32 & 25 & 81  \\ %91, 83
     
    \midrule
  \multirow{8}{*}{\begin{sideways}\textit{EORSSD}~\cite{2021DAFNet}\end{sideways}}
    & $S_{\alpha}\uparrow$    & .5992 & .5971 & .6420 & .7101 & .6007
    					    & .7868 & .8179 & .8184 & .8348 & .8207 & .8601 
					    & .8869 & .9040 & .9050 & .9114
					    & .5082 & .5798 & .5401 & .8630 & \textcolor{green}{\textbf{.9166}} & \textcolor{blue}{\textbf{.9197}} & \textcolor{red}{\textbf{.9290}} & \textcolor{red}{\textbf{.9290}} \\
    & max $F_{\beta}\uparrow$     & .3993 & .5407 & .5232 & .5884 & .3995
    					    & .6849 & .7446 & .7498 & .7454 & .7545 & .7880 
					    & .8347 & .8344 & .8523 & .8566
					    & .2765 & .3268 & .5176 & .7794 & .8614  & \textcolor{green}{\textbf{.8656}} & \textcolor{blue}{\textbf{.8720}} & \textcolor{red}{\textbf{.8837}} \\
    & mean $F_{\beta}\uparrow$     & .3686 & .4018 & .4597 & .5473 & .3685
    					    & .5801 & .6582 & .6302 & .6766 & .6406 & .6967 
					    & .7905 & .8174 & .8221 & .8228
					    & .2107 & .2692 & .2992 & .7328 & .7845 & \textcolor{green}{\textbf{.8239}} & \textcolor{blue}{\textbf{.8486}} &\textcolor{red}{\textbf{.8552}} \\
    & adp $F_{\beta}\uparrow$     & .3344 & .2658 & .4012 & .4081 & .3347
    					    & .4597 & .4933 & .4165 & .5471 & .4611 & .5379 
					    & .6723 & \textcolor{green}{\textbf{.7705}} & .7421 & .7109
					    & .1836 & .2007 & .2083 & .6284 & .6427 & .7066 & \textcolor{red}{\textbf{.7984}} & \textcolor{blue}{\textbf{.7969}} \\
    & max $E_{\xi}\uparrow$          & .6894 & .7861 & .7260 & .7697 & .6882
    					    & .9186 & .9140 & .9483 & .9266 & .9292 & .9570 
					    & .9524 & .9442 & .9556 & .9610
					    & .5982 & .6803 & .7744 & .9254 & \textcolor{red}{\textbf{.9861}} & .9646 & \textcolor{green}{\textbf{.9711}} & \textcolor{blue}{\textbf{.9727}} \\
    & mean $E_{\xi}\uparrow$          & .5943 & .6376 & .6594 & .7286 & .5946
    					    & .7631 & .8567 & .8294 & .8638 & .8193 & .8775 
					    & .9167 & .9346 & \textcolor{green}{\textbf{.9407}} & .9385
					    & .4886 & .5894 & .5197 & .8801 & .9291 & .9350 & \textcolor{blue}{\textbf{.9604}} & \textcolor{red}{\textbf{.9653}} \\
    & adp $E_{\xi}\uparrow$          & .5639 & .5192 & .6188 & .6416 & .5636
    					    & .6933 & .7261 & .6462 & .7738 & .6836 & .7566 
					    & .8647 & \textcolor{green}{\textbf{.9243}} & .9103 & .8909
					    & .4767 & .4890 & .5014 & .8445 & .8446 & .8897 & \textcolor{red}{\textbf{.9501}} & \textcolor{blue}{\textbf{.9450}} \\
    & $\mathcal{M}\downarrow$ & .1677 & .1088 & .1246 & .0771 & .1644
    					    & .0186 & .0168 & .0171 & .0160 & .0210 & .0110 
					    & .0102 & .0093 & .0106 & .0095 
					    & .2096 & .1057 & .1434 & .0146 & \textcolor{red}{\textbf{.0060}} & .0099 & \textcolor{green}{\textbf{.0084}} & \textcolor{blue}{\textbf{.0074}} \\
    \midrule
  \multirow{8}{*}{\begin{sideways}\textit{ORSSD}~\cite{2019LVNet}\end{sideways}}
    & $S_{\alpha}\uparrow$    & .6835 & .6197 & .7195 & .7640 & .6849
    					    & .8262 & .8259 & .8141 & .8613 & .8403 & .8721 
					    & .9026 & .9040 & .9050 & .9186
					    & .5366 & .6033 & .5312 & .8815 & .9191 & \textcolor{green}{\textbf{.9204}} & \textcolor{blue}{\textbf{.9366}} & \textcolor{red}{\textbf{.9437}} \\
    & max $F_{\beta}\uparrow$     & .5590 & .5257 & .6238 & .6692 & .5591
    					    & .7467 & .7619 & .7456 & .8131 & .7706 & .8332 
					    & .8687 & .8761 & .8735 & .8871
					    & .3471 & .3913 & .4417 & .8263 & \textcolor{green}{\textbf{.8928}} & .8842 & \textcolor{blue}{\textbf{.9002}} & \textcolor{red}{\textbf{.9149}} \\
    & mean $F_{\beta}\uparrow$     & .5125 & .4235 & .5747 & .6214 & .5126
    					    & .6962 & .6856 & .7383 & .7308 & .6999 & .7500 
					    & .8433 & .8574 & .8502 & \textcolor{green}{\textbf{.8679}}
					    & .2717 & .3454 & .2684 & .7995 & .8511 & .8566 & \textcolor{blue}{\textbf{.8856}} & \textcolor{red}{\textbf{.8971}} \\
    & adp $F_{\beta}\uparrow$     & .4874 & .3722 & .5657 & .5568 & .4876
    					    & .6206 & .5730 & .7379 & .6722 & .6166 & .6452 
					    & .7861 & \textcolor{green}{\textbf{.8251}} & .8068 & .8229
					    & .2633 & .3108 & .2496 & .7506 & .7876 & .8022 & \textcolor{blue}{\textbf{.8617}} & \textcolor{red}{\textbf{.8806}} \\
    & max $E_{\xi}\uparrow$          & .7649 & .7719 & .7912 & .8230 & .7651
    					    & .8860 & .9130 & .8913 & .9519 & .9343 & .9731
					    & .9509 & .9545 & .9601 & .9664
					    & .6514 & .7064 & .7402 & .9456 & \textcolor{blue}{\textbf{.9771}} & .9623 & \textcolor{green}{\textbf{.9737}} & \textcolor{red}{\textbf{.9796}} \\
    & mean $E_{\xi}\uparrow$          & .7017 & .6495 & .7337 & .7745 & .7021
    					    & .8362 & .8298 & .8681 & .8553 & .8650 & .9013 
					    & .9341 & .9454 & .9482 & .9538
					    & .5352 & .6417 & .4920 & .9259 & \textcolor{green}{\textbf{.9539}} & .9415 & \textcolor{blue}{\textbf{.9671}} & \textcolor{red}{\textbf{.9754}} \\
    & adp $E_{\xi}\uparrow$          & .6949 & .6291 & .7532 & .7682 & .6950
    					    & .8085 & .7678 & .8887 & .8504 & .8124 & .8226 
					    & .9205 & .9423 & .9335 & \textcolor{green}{\textbf{.9428}} 
					    & .5826 & .5996 & .5676 & .9195 & .9360 & .9328 & \textcolor{blue}{\textbf{.9663}} & \textcolor{red}{\textbf{.9721}} \\
    & $\mathcal{M}\downarrow$ & .1324 & .1309 & .1041 & .0715 & .1277
    					    & .0363 & .0382 & .0399 & .0243 & .0358 & .0216 
					    & .0168 & .0144 & .0165 & .0137
					    & .2151 & .1267 & .1854 & .0207 & \textcolor{green}{\textbf{.0113}} & .0163 & \textcolor{blue}{\textbf{.0109}} & \textcolor{red}{\textbf{.0088}} \\
%    \midrule
%     \multicolumn{2}{c|}{AveRanking} & 0.786 & 0.806 & 0.787 & 0.819 & xx
%    					   & 0.846 & 0.809 & 0.847 & 0.887 & xx & xx
%					   & 0.912 & 0.925 & 0.927 & 0.923 & xx & \textcolor{red}{0.942} \\
    \bottomrule
  \hline
  \end{tabular}}
\end{table*}
%%%
%%%

\section{Experimental Results}
\label{sec:exp}

\subsection{Experimental Protocol}
\label{sec:ExpProtocol}
\textit{1) Datasets.}
We evaluate the proposed method on two recently proposed datasets for RSI-SOD.

\textbf{ORSSD}~\cite{2019LVNet} is the first publicly available dataset for RSI-SOD,
collected from the Google Earth and some existing RSI datasets.
It contains 800 optical RSIs and provides corresponding pixel-wise annotation for each image.
Among these optical RSIs, 600 images are used as training set and the remaining 200 images as testing set.

\textbf{EORSSD}~\cite{2021DAFNet} is the largest public dataset for RSI-SOD.
It extends the original ORSSD dataset to 2,000 images with corresponding pixel-wise GTs.
Among these, 1,400 images are used as training set and 600 images as testing set.

\textit{2) Network Training Details.}
We implement the proposed ACCoNet by PyTorch~\cite{PyTorch} with an NVIDIA Titan X GPU.
In the training and testing phases, the input optical RSIs are resized into $256\times256$.
We adopt the parameters of the pre-trained VGG-16 model~\cite{2014VGG16ICLR} to initialize the parameters of the encoder network in our ACCoNet,
while the parameters of all other newly added layers are initialized by the normal distribution~\cite{InitialWei}. 
We set the initial learning rate to $1e^{-4}$, and it will be divided by 10 after 30 epochs.
Due to the limitation of GPU memory, we set the batch size to 6.
We use the Adam optimizer~\cite{Adam} for network optimization.
For data augmentation, we adopt the flipping and rotation, producing seven additional variants of the original training data.
Specifically, on the EORSSD dataset~\cite{2021DAFNet}, we train our ACCoNet with 11,200 augmented pairs for 39 epochs.
On the ORSSD dataset~\cite{2019LVNet}, we train our ACCoNet with 4,800 augmented pairs for 54 epochs.

\textit{3) Evaluation Metrics.}
We adopt nine widely used evaluation metrics, including
S-measure ($S_{\alpha}$, $\alpha$ = 0.5)~\cite{Fan2017Smeasure},
maximum, mean and adaptive F-measure ($F_{\beta}$, $\beta^2$ = 0.3)~\cite{Fmeasure},
maximum, mean and adaptive E-measure ($E_{\xi}$)~\cite{Fan2018Emeasure},
Mean Absolute Error (MAE, $\mathcal{M}$) and Precision-Recall (PR) curve,
to comprehensively measure the performance of our ACCoNet and other compared methods. 
Specifically, $\textbf{S-measure}$ simultaneously measures the region-aware and object-aware structural similarity.
$\textbf{F-measure}$ is the weighted harmonic mean of precision and recall, and we pay more attention to precision in the paper.
$\textbf{E-measure}$ jointly considers the local pixel-level match information and the global image-level statistics.
$\textbf{MAE}$ evaluates the average pixel-level errors.
$\textbf{PR curve}$ presents the correlation between precision and recall.
The evaluation tool\footnote{http://dpfan.net/d3netbenchmark/} provided by Fan \etal\cite{Fan2019D3Net} is adopted by us for convenient evaluation.

%%%
%%%
\begin{figure}
\centering
\footnotesize
  \begin{overpic}[width=1\columnwidth]{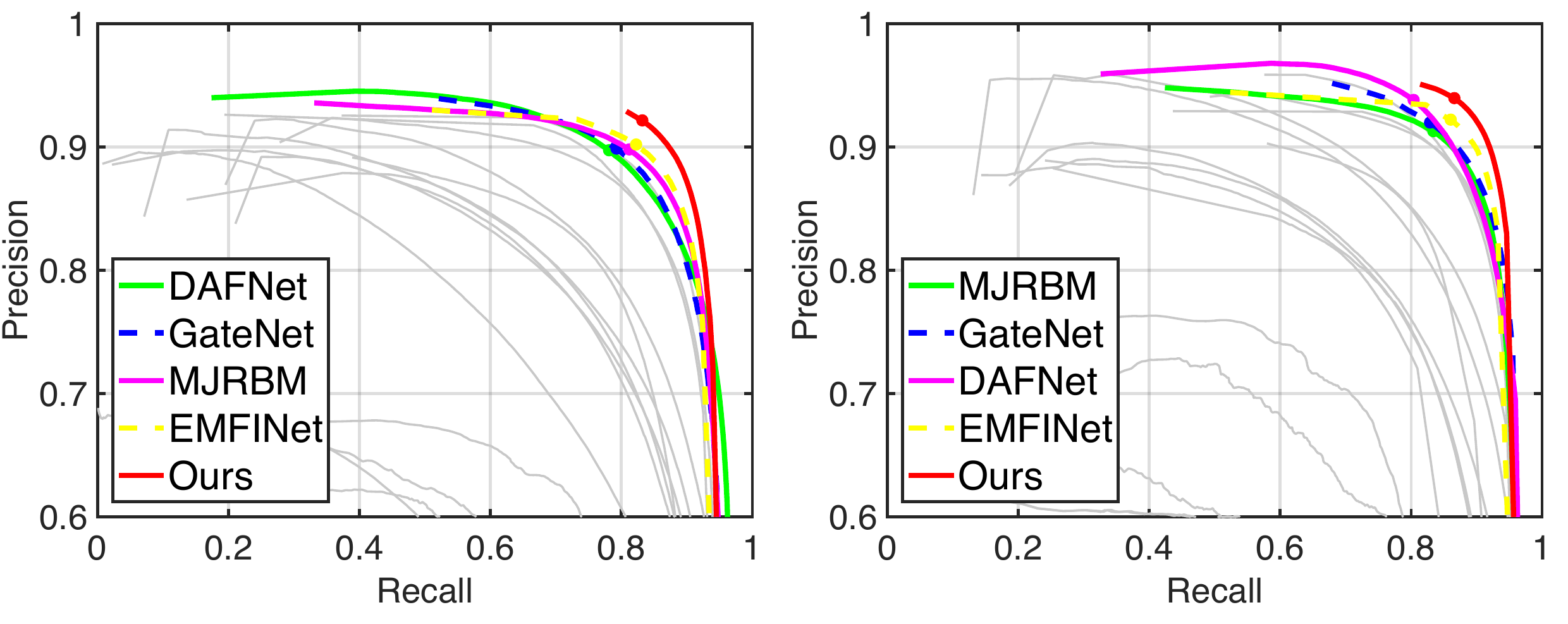}
    \put(15.6,-2.3){ (a) EORSSD~\cite{2021DAFNet} }
    \put(66.6,-2.3){ (b) ORSSD~\cite{2019LVNet} }   
  \end{overpic}
\caption{
Quantitative performance comparison on PR curve in two datasets.
The top five methods are shown in color.
Please zoom-in for details.
}
\label{PR_comparison}
\end{figure}
%%%
%%%

\subsection{Comparison with State-of-the-arts}
\textit{1) Comparison Methods.}
Following the two popular RSI-SOD benchmarks~\cite{2019LVNet,2021DAFNet}, we compare our method with 22 state-of-the-art NSI-SOD and RSI-SOD methods for a comprehensive evaluation.
Concretely, these compared methods include
five traditional NSI-SOD methods (RRWR~\cite{2015RRWR}, HDCT~\cite{2016HDCT}, DSG~\cite{2017DSG}, SMD~\cite{2017SMD}, and RCRR~\cite{2018RCRR}),
ten CNN-based NSI-SOD methods (DSS~\cite{2017DSS}, RADF~\cite{2018RADF}, R3Net~\cite{2018R3Net}, PFAN~\cite{2019PFAN}, PoolNet~\cite{2019PoolNet}, EGNet~\cite{2019EGNet}, GCPA~\cite{2020GCPA}, MINet~\cite{2020MINet}, ITSD~\cite{2020ITSD}, and GateNet~\cite{2020GateNet}),
three traditional RSI-SOD methods (VOS~\cite{2018VOS}, CMC~\cite{2019CMC}, and SMFF~\cite{2019SMFF}), and
four recent CNN-based RSI-SOD methods (LVNet~\cite{2019LVNet}, DAFNet~\cite{2021DAFNet}, MJRBM~\cite{2021MJRBM}, and EMFINet~\cite{2021EMFINet}).
Notably, except for GCPA~\cite{2020GCPA}, MINet~\cite{2020MINet}, ITSD~\cite{2020ITSD} and GateNet~\cite{2020GateNet}, the saliency maps of all the other compared methods are provided by Zhang \etal\cite{2021DAFNet}\footnote{https://github.com/rmcong/DAFNet\_TIP20} and/or by the authors.
Following~\cite{2019LVNet, 2021DAFNet}, we fine-tune GCPA~\cite{2020GCPA}, MINet~\cite{2020MINet}, ITSD~\cite{2020ITSD} and GateNet~\cite{2020GateNet} with their default hyperparameter settings using the same training data as our method on the two datasets.

\textit{2) Quantitative Comparison on EORSSD.}
We present the quantitative comparison of EORSSD~\cite{2021DAFNet} in terms of $S_{\alpha}$, $F_{\beta}$, $E_{\xi}$ and $\mathcal{M}$ in the upper part of Tab.~\ref{table:QuantitativeResults}.
Among the eight metrics in Tab.~\ref{table:QuantitativeResults}, our method ranks first in four metrics and second in other four metrics.
Overall, on the EORSSD dataset, our method performs the best among all compared methods.
EMFINet~\cite{2021EMFINet} is the best among the seven existing RSI-SOD methods, and GateNet~\cite{2020GateNet} is the best among existing NSI-SOD methods.
In comparison to EMFINet, our method performs marginally lower in terms of adp $F_{\beta}$ and adp $E_{\xi}$, but surpasses EMFINet by 1.17\% on max $F_{\beta}$.
Compared with GateNet, our method greatly outperforms it by 2.71\%, 3.24\%, 5.41\% and 8.60\% on max $F_{\beta}$, mean $F_{\beta}$, adp $E_{\xi}$ and adp $F_{\beta}$, respectively.
In addition, we show the PR curve in Fig.~\ref{PR_comparison}(a), and our method is better than all compared methods.

\textit{3) Quantitative Comparison on ORSSD.}
The quantitative comparison of ORSSD~\cite{2019LVNet} on eight metrics is shown at the bottom part of Tab.~\ref{table:QuantitativeResults}, and the PR curve is shown in Fig.~\ref{PR_comparison}(b).
Our method consistently outperforms all compared methods among all nine quantitative metrics.
Notably, compared with the second best method, the performance gain of our method reaches
1.89\% on adp $F_{\beta}$,
1.47\% on max $F_{\beta}$,
and 1.15\% on mean $F_{\beta}$.
Among all the compared methods, ours is the only method whose $\mathcal{M}$ is lower than 0.0100, \ie 0.0088.

According to the quantitative comparison on the two datasets, our method is the best method for RSI-SOD.
In addition, comparing the specialized RSI-SOD methods and the NSI-SOD methods in the same period, we can find that the specialized methods are better than the NSI-SOD methods, which indicates that the development of specialized methods is necessary and urgent.

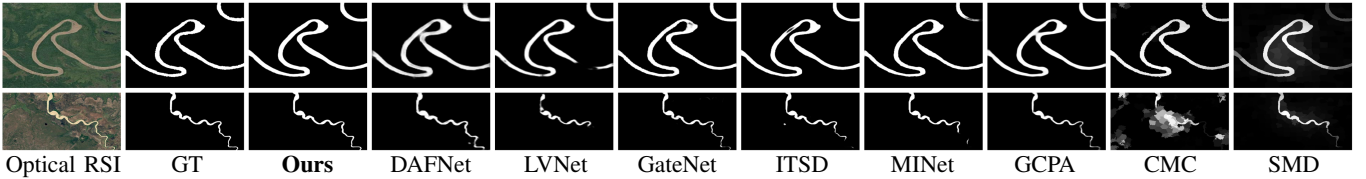
\begin{figure*}[t!]
    \centering
    \small
	\begin{overpic}[width=1\textwidth]{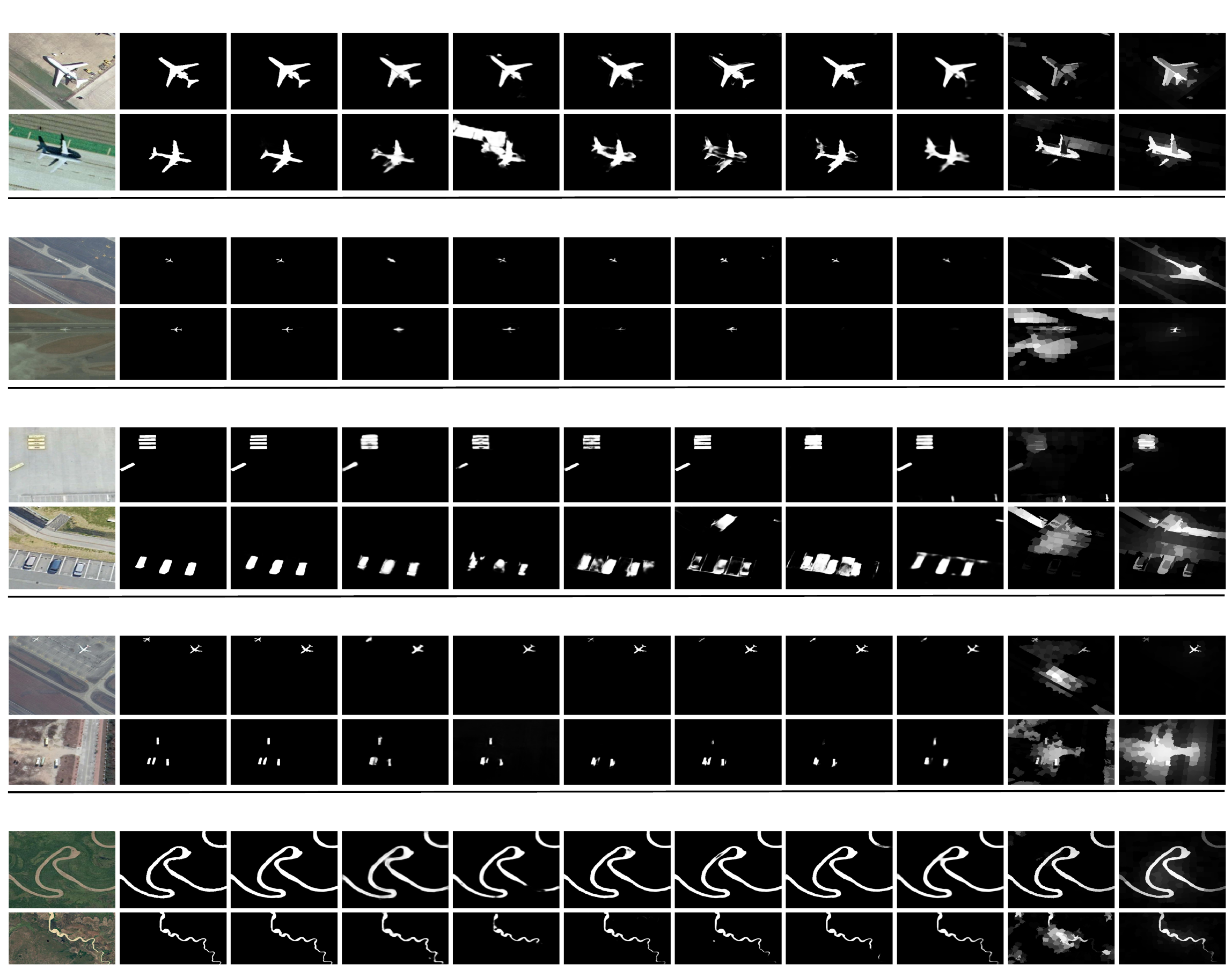}

    \put(0.25,77.3){ Object with shadows}
    \put(0.25,60.8){ Tiny object}
    \put(0.25,45.35){ Multiple objects}
    \put(0.25,28.45){ Multiple tiny objects}
    \put(0.25,12.55){ Irregular geometry structure}

    \put(0.25,-0.65){ Optical RSI}
    \put(12.4,-0.65){ GT}
    \put(20.5,-0.65){ \textbf{Ours}}
    \put(28.45,-0.65){ DAFNet}
    \put(38.25,-0.65){ LVNet}
    \put(46.58,-0.65){ GateNet}
    \put(56.7,-0.65){ ITSD}
    \put(65.15,-0.65){ MINet}
    \put(74.2,-0.65){ GCPA}
    \put(83.7,-0.65){ CMC}
    \put(92.8,-0.65){ SMD} %\scriptsize
    
    \end{overpic}
	\caption{Visual comparisons with eight representative state-of-the-art methods,
	including two CNN-based RSI-SOD methods (DAFNet~\cite{2021DAFNet} and LVNet~\cite{2019LVNet}),
	four CNN-based NSI-SOD methods (GateNet~\cite{2020GateNet}, ITSD~\cite{2020ITSD}, MINet~\cite{2020MINet} and GCPA~\cite{2020GCPA}),
	one traditional RSI-SOD method (CMC~\cite{2019CMC}),
	and one traditional NSI-SOD method (SMD~\cite{2017SMD}).
	Please zoom-in for the best view, especially for tiny object and multiple tiny objects.
    }
    \label{fig:VisualExample}
\end{figure*}

\textit{4) Visual Comparison.}
We show some qualitative results in Fig.~\ref{fig:VisualExample}, including several representative and challenging scenes of optical RSIs, such as object with shadows, tiny object, multiple objects, multiple tiny objects, and irregular geometry structure.

For the first scene, shadows are usually connected with salient objects, which will interfere with the detection and highlight inaccurate regions on the saliency map.
We can clearly observe that in the second example, LVNet, GateNet, ITSD, CMC and SMD are in this dilemma, but our method can highlight the plane more accurately.

The second scene is unique to optical RSIs and is different from the scene with the small object in NSIs.
In this scene, optical RSIs contain much smaller object, \ie the tiny object.
Such extreme scene invalidates traditional methods and two CNN-based NSI-SOD methods, \ie CMC, SMD, MINet and GCPA.
The first two methods detect wrong objects in the first example, and the latter two methods fail to detect any objects in the second example.
Besides, other methods can only roughly determine the location of the tiny object but the details cannot be described well.
Our method can capture the tiny object with fine details.

%%%
%%%
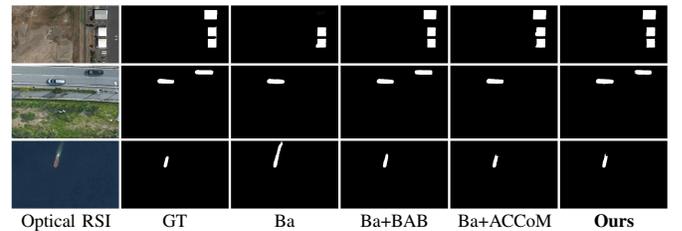
\begin{figure}
\centering
\footnotesize
  \begin{overpic}[width=1\columnwidth]{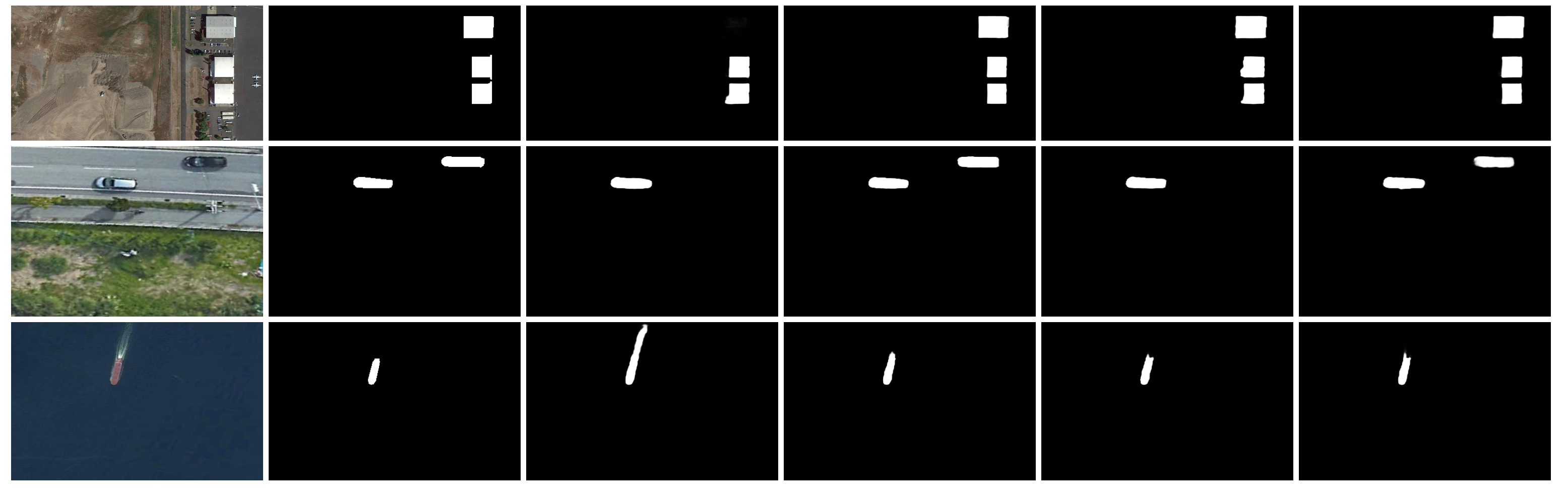}
    \put(0.9,-2){ \scriptsize{Optical RSI}}
    \put(22.1,-2){ \scriptsize{GT}}
    \put(38.9,-2){ \scriptsize{Ba}}
    \put(51.8,-2){ \scriptsize{Ba+BAB}}
    \put(66.45,-2){ \scriptsize{Ba+ACCoM}}
    \put(86.9,-2){ \scriptsize{\textbf{Ours}} }   
  \end{overpic}
\caption{
Visual examples of ablation studies.
``Ba" represents the basic encoder-decoder network.
Please zoom-in for details.
}
\label{AblationStudy1}
\end{figure}
%%%
%%%

%
Scene with multiple objects has always been the difficulty of the SOD task.
In the first example, MINet misses an object.
Although other methods detect all objects, objects are incomplete.
In the second example, due to the complexity of the scene, GateNet, ITSD, MINet, CMC and SMD incorrectly detect more regions.
On the contrary, our method locates all objects finely without any redundant regions.

The fourth scene is a combination of the second and third scenes, which puts forward higher requirements for the SOD method.
All representative compared methods appear to miss the detection of some objects, while our method distinguishes all tiny objects.

The last scene refers specifically to the river.
Rivers often have very complex irregular geometric structures and span the entire image.
They have different widths in different positions, which is not friendly to some methods, causing LVNet, ITSD and MINet to detect only part of the river.
Thanks to our method thoroughly explores the contextual information in both encoder and decoder, which is particularly advantageous for variable object scales, object shapes and object quantities in optical RSIs, our method can overcome the above common and complex scenes in optical RSIs.

\textit{5) Speed Comparison.}
In Tab.~\ref{table:QuantitativeResults}, we report the speed of 15 compared methods and ours\footnote{The speed of RRWR, HDCT, DSG, RCRR, DSS, RADF, R3Net and LVNet are borrowed from~\cite{2019LVNet}, the speed of GCPA, MINet, ITSD, GateNet and MJRBM is obtained by our test, and the speed of DAFNet and EFMINet is obtained from original papers.}.
Our method reaches a fast processing speed of 81 \textit{fps} on a GPU, which ranks first among 16 compared methods and is more than three times that of the second best method EFMINet (\ie 25 \textit{fps}).
Based on the above comprehensive comparison, our method shows remarkable detection accuracy and astonishing speed.

%%%
%%%
\begin{table}[!t]
\centering
%\small
\caption{Ablation analyses on measuring the overall contributions of ACCoM and BAB in ACCoNet.
  Baseline is the encoder-decoder network.
  The best result in each column is \textbf{bold}.
  }
\label{table:AblationStudyAll}
%\footnotesize
\renewcommand{\arraystretch}{1.5}
\renewcommand{\tabcolsep}{0.6mm}
%\small
\begin{tabular}{c|ccc||ccc|ccc}
\toprule

 \multirow{2}{*}{No.} & \multirow{2}{*}{Baseline} & \multirow{2}{*}{ACCoM} & \multirow{2}{*}{BAB}  
 & \multicolumn{3}{c|}{EORSSD~\cite{2021DAFNet}} 
 & \multicolumn{3}{c}{ORSSD~\cite{2019LVNet}} \\
 
 \cline{5-10}
    & & & 
    & max$F_{\beta}$ & $\mathcal{M}$ & max$E_{\xi}$
    & max$F_{\beta}$ & $\mathcal{M}$ & max$E_{\xi}$ \\
\midrule
1 & \Checkmark   &                     &                     & .8642 & .0093 & .9547 & .8832 & .0138 & .9566  \\ 
2 &  \Checkmark & \Checkmark &                     & .8819 & .0076 & .9673 & .9117 & .0098 & .9766  \\
3 &  \Checkmark &                     & \Checkmark & .8777 & .0086 & .9655 & .8987 & .0131 & .9661   \\
\hline
4 &  \Checkmark & \Checkmark & \Checkmark & \bf{.8837} & \bf{.0074} & \bf{.9727} & \bf{.9149} & \bf{.0088} & \bf{.9796} \\
\toprule
\end{tabular}
\end{table}
%%%
%%%

\subsection{Ablation Studies}
\label{Ablation Studies}
In this subsection, we conduct thorough ablation studies on EORSSD~\cite{2021DAFNet} and ORSSD~\cite{2019LVNet} to investigate the impact of the two vital components in our method.
Specifically, we analyze
1) the overall contributions of ACCoM and BAB in ACCoNet,
2) the effectiveness of two types of branches in ACCoM,
3) the rationality of the dilated convolution based bifurcations in BAB,
4) the complementarity between BCE and IoU in loss function, and
5) the flexibility of our method.
For each variant, we strictly modify only one part at a time and retrain the variant on the two datasets using the same training settings as in Sec.~\ref{sec:ExpProtocol}.

\textbf{1. The overall contributions of ACCoM and BAB in ACCoNet.}
As shown in Tab.~\ref{table:AblationStudyAll}, to measure the overall contributions of the proposed ACCoM and BAB to ACCoNet, we offer three variants:
1) the encoder-decoder network (\ie ``Baseline"),
2) the baseline network with only ACCoMs (\ie ``Baseline+ACCoM"), and
3) the baseline network with only BABs (\ie ``Baseline+ACCoM").
Besides, the complete ACCoNet is ``Baseline+ACCoM+BAB".
We report the quantitative results in Tab.~\ref{table:AblationStudyAll}.

On the EORSSD dataset, we observe that ``Baseline" only achieves 86.42\% on max $F_{\beta}$, 0.0093 on $\mathcal{M}$ and 95.47\% on max $E_{\xi}$.
ACCoM increases ``Baseline" by 1.76\%, 0.0017 and 1.26\% on these three metrics respectively, while BAB increases ``Baseline" by 1.35\%, 0.0007 and 1.08\% on these three metrics respectively.
With the joint cooperation of ACCoM and BAB, our complete ACCoNet improves ``Baseline" by 1.95\%, 0.0017 and 1.80\% on these three metrics respectively.
The trends on the ORSSD dataset are the same as that on the EORSSD dataset.
Notably, our complete ACCoNet improves ``Baseline" by 3.17\%, 0.0050 and 2.30\% on max $F_{\beta}$, $\mathcal{M}$ and max $E_{\xi}$, respectively, which more clearly validates the effectiveness of each proposed module.

\begin{table}[t!]
  \centering
  \renewcommand{\arraystretch}{1.5}
  \renewcommand{\tabcolsep}{0.7mm}
  \caption{
  Ablation results on confirming the effectiveness of two types of branches in ACCoM and the rationality of the dilated convolution based bifurcations in BAB.
  The best result in each column is \textbf{bold}.
  }
  \label{table:AC2M_BAB}
 
\begin{tabular}{c||ccc|ccc}
\bottomrule[1pt]
 \multirow{2}{*}{Models} 
 & \multicolumn{3}{c|}{EORSSD~\cite{2021DAFNet}} 
 & \multicolumn{3}{c}{ORSSD~\cite{2019LVNet}}\\
\cline{2-7}
    & max$F_{\beta}\uparrow$ & $\mathcal{M}\downarrow$ & max$E_{\xi}\uparrow$
    & max$F_{\beta}\uparrow$ & $\mathcal{M}\downarrow$ & max$E_{\xi}\uparrow$ \\ 
%\midrule[1pt]
\hline
\hline

ACCoNet (\textbf{Ours}) & \bf{.8837} & \bf{.0074} & \bf{.9727} 
		    & \bf{.9149} & \bf{.0088} & \bf{.9796}  \\
\hline

\textit{w/o LB}  & .8800 & .0079 & .9681 %w/o Intra (glo)
         	         & .9029 & .0113 & .9691  \\			
\textit{w/o AB}  & .8830 & .0075 & .9704 %w/o Inter (loc)
		         & .9072 & .0108 & .9739 \\
\hline
\textit{w/ DC}  & .8831 & .0075 & \bf{.9727}  %w/o Intra (glo)
         	         & .9136 & .0093 & .9790  \\			
\textit{w/ NC}  & .8834 & \bf{.0074} & .9716 %w/o Inter (loc)
		         & .9144 & .0090 & .9783 \\
\toprule[1pt]
\multicolumn{7}{l}{\scriptsize{\textit{w/o LB}: ACCoM without local branch. \textit{w/o AB}: ACCoM without adjacent branches. } } \\
%\multicolumn{7}{l}{\scriptsize{\textit{w/o AB}: ACCoM without adjacent branches.}} \\
\multicolumn{7}{l}{\scriptsize{\textit{w/ DC}: two bifurcations of BAB are direct connection operations. }} \\
\multicolumn{7}{l}{\scriptsize{\textit{w/ NC}: two bifurcations of BAB are normal convolutional layers.}} \\
\end{tabular}
\end{table}
Additionally, we show saliency maps of these three variants and our method in Fig.~\ref{AblationStudy1}.
In the first and second examples, ``Ba" (\ie ``Baseline") misses an object.
In the first example, both ACCoM and BAB complete the missing object.
Differently, in the second example, only BAB completes the missing object.
This means that as long as ACCoM or BAB can complete the missing object, the complete ACCoNet (\ie ``Ours") can get accurate saliency maps.
In the third example, ``Ba" mistakenly highlights the background region.
BAB suppresses part of the background and ACCoM suppresses more background, resulting in a satisfactory saliency map of ``Ours".
The above quantitative and qualitative analysis confirms that both ACCoM and BAB are important for ACCoNet, and the contextual information explored by these two modules is conducive to the detection of salient objects in optical RSIs.

%%%
%%%
\begin{figure}
\centering
\footnotesize
  \begin{overpic}[width=1\columnwidth]{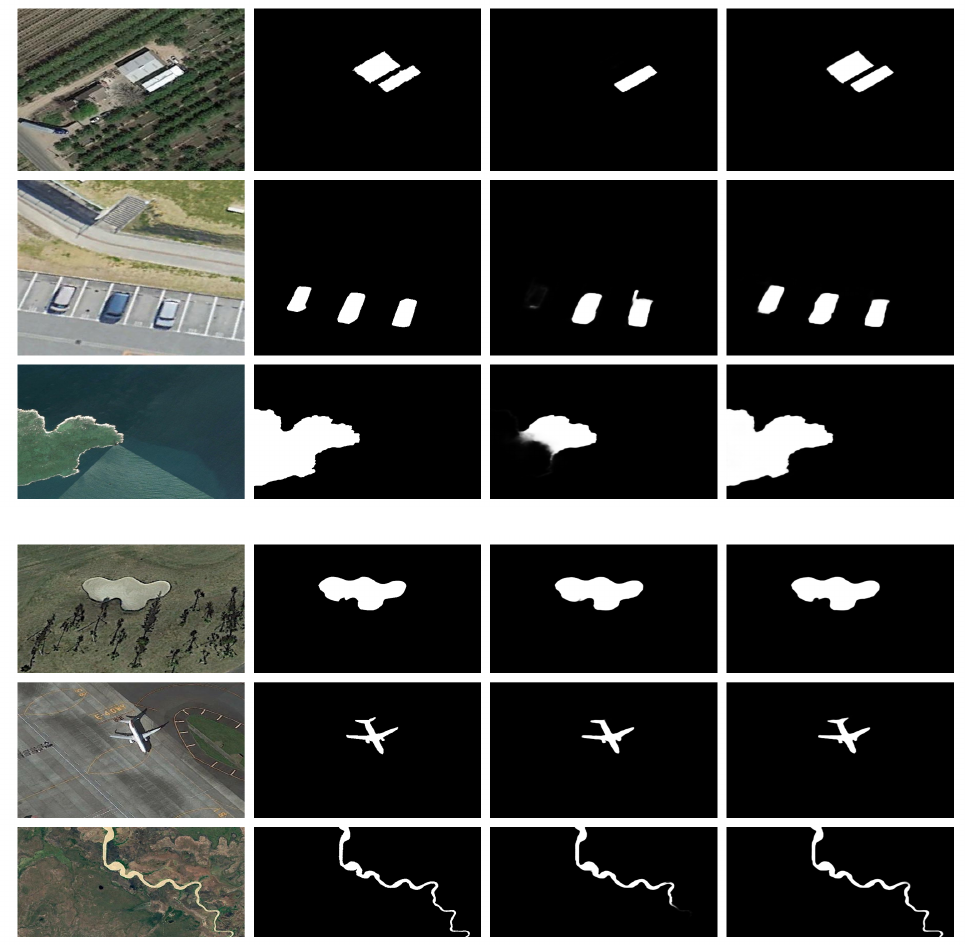}
    \put(4.8,43.5){ {Optical RSI}}
    \put(34.8,43.5){ {GT}}
    \put(56.3,43.5){ {\textit{w/o LB}}}
    \put(82.2,43.5){ {\textbf{Ours}} }     
  
    \put(4.8,-1.9){ {Optical RSI}}
    \put(34.8,-1.9){ {GT}}
    \put(56.3,-1.9){ {\textit{w/o AB}}}
    \put(82.2,-1.9){ {\textbf{Ours}} }   
  \end{overpic}
\caption{
Visual examples of two variants, \textit{w/o LB} and \textit{w/o AB}.
}
\label{Fig:LBAB}
\end{figure}
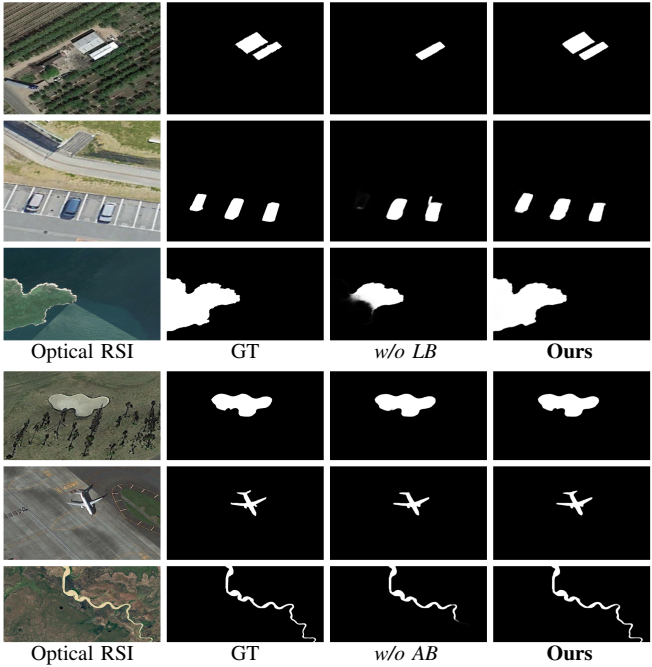
%%%
%%%

\textbf{2. The effectiveness of two types of branches in ACCoM.}
To investigate the effectiveness of two types of branches in ACCoM, we provide two variants:
1) removing the local branch in ACCoM (\ie \textit{w/o LB}) and
2) removing the adjacent branches in ACCoM (\ie \textit{w/o AB}).
The ablation results are reported in the third and fourth rows of Tab.~\ref{table:AC2M_BAB}.

We discover that the performances of \textit{w/o LB} and \textit{w/o AB} are worse than ours, which demonstrates that these two types of branches are effective.
Concretely, on the ORSSD dataset, the performance of \textit{w/o LB} is degraded,
\eg max $F_{\beta}$: 91.49\% $\rightarrow$ 90.29\%, $\mathcal{M}$: 0.0088 $\rightarrow$ 0.0113, max $E_{\xi}$: 97.96\% $\rightarrow$ 96.91\%,
while the performance of \textit{w/o AB} drops slightly,
\eg max $F_{\beta}$: 91.49\% $\rightarrow$ 90.72\%, $\mathcal{M}$: 0.0088 $\rightarrow$ 0.0108, max $E_{\xi}$: 97.96\% $\rightarrow$ 97.39\%.
The same trend is observed on the EORSSD dataset.
The reason is that the feature modulation of adjacent branches is based on $ \boldsymbol{f}^{t}_\textrm{c}$, which belongs to the local branch.
If we remove the local branch, the global assistance provided by two adjacent features will act on $ \boldsymbol{f}^{t}_\textrm{e} $, which cannot exert the maximum effect of global assistance.
Thus, we conclude that the local branch is the core of ACCoM.

Specifically, in Fig.~\ref{Fig:LBAB}, we show saliency maps of these two variants and our complete method to visually evaluate the role of the local branch and the adjacent branches.
As shown in the first three examples of Fig.~\ref{Fig:LBAB}, the saliency maps of \textit{w/o LB} miss objects in the case of multiple salient objects (the first two examples), and cannot detect the complete object in the case of large salient object (the third one).
This is because after removing the local branch, the location information of salient objects will be reduced, resulting in two types of missed detections.
Differently, for the saliency maps of \textit{w/o AB}, the salient objects are basically located accurately, but the details are not perfectly outlined, such as the regions occluded by the tree (the fourth one), the airplane tail (the fifth one), and the slender river (the last one).
After removing the adjacent branches, the cross-level contextual complementary information is discarded, causing the damage of the salient object details.
In summary, the local branch is good for scenes with multiple salient objects and large salient object, while the adjacent branches are good for scenes containing salient objects with fine details. 

\textbf{3. The rationality of the dilated convolution based bifurcations in BAB.}
To validate the rationality of the dilated convolution based bifurcations in BAB, we conduct two variants:
1) replacing dilated convolutions by direct connection operations (\ie \textit{w/ DC}) and
2) replacing dilated convolutions by normal convolutional layers (\ie \textit{w/ NC}).
The ablation results are reported in the last two rows of Tab.~\ref{table:AC2M_BAB}.

In general, we find that the performance gap between these two variants and our original BAB is small.
However, with direct connection operations, BAB cannot fully demonstrate its ability to capture contextual information, which leads to performance degradation,
\eg max $F_{\beta}$: 88.31\% (\textit{w/ DC}) \textit{v.s.} 88.37\% (Ours) on the EORSSD and 91.36\% (\textit{w/ DC}) \textit{v.s.} 91.49\% (Ours) on the ORSSD.
The normal convolutional layers slightly improve the ability of BAB compared to direct connection operations,
\eg max $F_{\beta}$: 88.31\% (\textit{w/ DC}) $\rightarrow$ 88.34\% (\textit{w/ NC}) on the EORSSD and 91.36\% (\textit{w/ DC}) $\rightarrow$ 91.44\% (\textit{w/ NC}) on the ORSSD.
In summary, the dilated convolution based bifurcations can capture better various contextual information with different receptive fields in the decoder.

%%%
%%%
\begin{table}[!t]
\centering
%\small
\caption{Ablation study on evaluating the complementarity between BCE and IoU in loss function.
  The best result in each column is \textbf{bold}.
  }
\label{AblationStudyLoss}
%\small
\renewcommand{\arraystretch}{1.5}
\renewcommand{\tabcolsep}{0.7mm}
%\small
\begin{tabular}{c|cc||ccc|ccc}
\toprule

 \multirow{2}{*}{No.} & \multirow{2}{*}{BCE} & \multirow{2}{*}{IoU}
 & \multicolumn{3}{c|}{EORSSD~\cite{2021DAFNet}} 
 & \multicolumn{3}{c}{ORSSD~\cite{2019LVNet}} \\
 
 \cline{4-9}
    & & 
    & max$F_{\beta}\uparrow$ & $\mathcal{M}\downarrow$ & max$E_{\xi}\uparrow$
    & max$F_{\beta}\uparrow$ & $\mathcal{M}\downarrow$ & max$E_{\xi}\uparrow$ \\ 
\midrule
1 & \Checkmark   &                     & .8731 & .0085 & .9666 & .9018 & .0117 & .9703  \\ 
2 &  			   & \Checkmark & .8801 & .0081 & .9711 & .9027 & .0105 & .9747  \\ 
\hline
3 &  \Checkmark & \Checkmark  & \bf{.8837} & \bf{.0074} & \bf{.9727} & \bf{.9149} & \bf{.0088} & \bf{.9796} \\
\toprule
\end{tabular}
\end{table}
%%%
%%%

%%

\textbf{4. The complementarity between BCE and IoU in loss function.}
To prove the complementarity between BCE and IoU in loss function, we provide two variants:
1) training our method with only BCE loss and
2) training our method with only IoU loss.
We report the quantitative results in Tab.~\ref{AblationStudyLoss}.

As shown in Tab.~\ref{AblationStudyLoss}, training our ACCoNet with only BCE loss or IoU loss can achieve promising performance, but the performance of these two variants is worse than that of our complete loss function.
This is because BCE loss is a pixel-level supervision, and IoU loss is a map-level supervision.
The two losses train the network from different aspects, and they can complement each other.
Combining the two losses to train our method together is conducive to keeping the completeness of salient objects.
This composite loss function is popular in the field of SOD~\cite{19BASNet,21HAINet,2021EMFINet}.

\begin{table}[t!]
  \centering
  \renewcommand{\arraystretch}{1.5}
  \renewcommand{\tabcolsep}{0.7mm}
  \caption{
  Performance on different encoder backbones of our ACCoNet.
  }
  \label{table:backbones}
 
\begin{tabular}{c||ccc|ccc}
\bottomrule[1pt]
 \multirow{2}{*}{Models}
 & \multicolumn{3}{c|}{EORSSD~\cite{2021DAFNet}}
 & \multicolumn{3}{c}{ORSSD~\cite{2019LVNet}} \\
\cline{2-7}
    & max$F_{\beta}\uparrow$ & $\mathcal{M}\downarrow$ & max$E_{\xi}\uparrow$
    & max$F_{\beta}\uparrow$ & $\mathcal{M}\downarrow$ & max$E_{\xi}\uparrow$ \\ 
%\midrule[1pt]
\hline
\hline

ACCoNet-VGG     & .8837 & .0074 & .9727 & .9149 & .0088 & .9796  \\
ACCoNet-ResNet & .8821 & .0067 & .9759 & .9149 & .0087 & .9819  \\

\toprule[1pt]
\end{tabular}
\end{table}

\textbf{5. The flexibility of our method.}
To demonstrate the flexibility of our method, we provide a variant, namely ACCoNet-ResNet, which adopts ResNet-50~\cite{2016ResNet} as the encoder backbone, and report the performance in Tab.~\ref{table:backbones}.
As shown in Tab.~\ref{table:backbones}, with the more powerful encoder backbone ResNet-50, the performance of ACCoNet-ResNet is improved on most evaluation metrics as compared with our original method, \ie ACCoNet-VGG in Tab.~\ref{table:backbones}, whose encoder backbone is VGG-16.
We can conclude that our method shows strong adaptability to different encoder backbones.

\section{Conclusion}
\label{sec:con}
In this paper, we investigate the contextual knowledge in an encoder-decoder architecture and proposed an effective ACCoNet for RSI-SOD.
We believe that the contextual information is beneficial to tackle variable object scales, object shapes and object quantities in RSI-SOD.
In the encoder, we propose the Adjacent Context Coordination Module (ACCoM) to coordinate the adjacent features (\ie the current, previous and subsequent features) and explore adjacent information for salient regions activation.
In the decoder, we propose the Bifurcation-Aggregation Block (BAB) to capture the multi-scale contents for salient regions inference.
Both ACCoMs and BABs learn contextual information to improve the representation of salient objects.
In particular, we employ the deep supervision with hybrid losses to stabilize the network training.
Extensive experiments, including quantitative, visual and speed comparisons and ablation studies, demonstrate that the proposed method is superior to 22 relevant state-of-the-art methods, and the two proposed modules contribute significantly to performance.

% \section*{Acknowledgment}


\ifCLASSOPTIONcaptionsoff
  \newpage
\fi

\bibliographystyle{IEEEtran}
\bibliography{ORSIref}



% \end{thebibliography}


\iffalse
\begin{IEEEbiography}{Michael Shell}
Biography text here.
\end{IEEEbiography}

\begin{IEEEbiographynophoto}{John Doe}
Biography text here.
\end{IEEEbiographynophoto}

%\newpage

\begin{IEEEbiographynophoto}{Jane Doe}
Biography text here.
\end{IEEEbiographynophoto}
\fi

\end{document}